\documentclass[10pt]{article}

\usepackage{fullpage}
\usepackage{fancyhdr}
\usepackage{setspace}
\usepackage{parskip}
\usepackage{pgfplots}
\usepackage{titlesec}
\usepackage[section]{placeins}
\usepackage{xcolor}
\usepackage{breakcites}
\usepackage{lineno}
\usepackage{hyphenat}
\usepackage{subcaption}
\usepgfplotslibrary{external}
\usepackage[toc,page]{appendix}
\usepackage{tcolorbox}
\usepackage{listings}
\usepackage{chngcntr}
\usepackage{makecell}
\pgfplotsset{compat=1.18}

\PassOptionsToPackage{hyphens}{url}
\usepackage[colorlinks = true,
            linkcolor = blue,
            urlcolor  = blue,
            citecolor = blue,
            anchorcolor = blue]{hyperref}
\usepackage{etoolbox}

\usepackage{natbib}
\bibpunct{(}{)}{,}{a}{}{;}
\renewenvironment{abstract}
  {{\bfseries\noindent{\abstractname}\par\nobreak}\footnotesize}
  {\bigskip}

\newenvironment{keywords}
  {{\bfseries\noindent Keywords:\par\nobreak}\footnotesize}
  {\bigskip}

\newcommand\blfootnote[1]{%
  \begingroup
  \renewcommand\thefootnote{}\footnote{#1}%
  \addtocounter{footnote}{-1}%
  \endgroup
}

\titlespacing{\section}{0pt}{*3}{*1}
\titlespacing{\subsection}{0pt}{*2}{*0.5}
\titlespacing{\subsubsection}{0pt}{*1.5}{0pt}

\definecolor{titleblue}{HTML}{174A6E}

\usepackage{authblk}
\usepackage{graphicx}
\usepackage[space]{grffile}
\usepackage{latexsym}
\usepackage{textcomp}
\usepackage{longtable}
\usepackage{tabulary}
\usepackage{booktabs,array,multirow}
\usepackage{amsfonts,amsmath,amssymb}
\providecommand\citet{\cite}
\providecommand\citep{\cite}

\newif\iflatexml\latexmlfalse

\AtBeginDocument{\DeclareGraphicsExtensions{.pdf,.PDF,.eps,.EPS,.png,.PNG,.tif,.TIF,.jpg,.JPG,.jpeg,.JPEG}}

\usepackage[utf8]{inputenc}
\usepackage[english]{babel}
\usepackage{float}

\newcommand{\runningauthor}{\small\textsc{O. Rabasseda et al.}}
\newcommand{\runningtitle}{\small\textit{Benchmarking MLLMs for Zero-Shot HOI}}
\fancypagestyle{plain}{%
  \fancyhf{}
  \fancyhead[L]{\runningauthor}
  \fancyhead[R]{\runningtitle}
  \fancyfoot[C]{\thepage}
  
}

\begin{document}

\title{\bfseries\LARGE Benchmarking Multimodal Large Language Models for Zero-Shot Human-Object Interaction\\[0.65em]
       {\rule{0.28\textwidth}{1.2pt}}}
\date{}

\author[1, 2, *]{Oriol Rabasseda}
\author[1]{Zenjie Li}
\author[1, 3]{Kamal Nasrollahi}
\author[2]{Sergio Escalera}
\affil[1]{Milestone Systems A/S, Banemarksvej 50, 2605 Broendby, Denmark}
\affil[2]{Universitat de Barcelona, Gran Via de les Corts Catalanes 585, 08007, Barcelona, Spain}
\affil[3]{Aalborg University, Rendsburggade 14, 9000 Aalborg, Denmark}

\renewcommand{\Authfont}{\normalsize\bfseries}
\renewcommand{\Affilfont}{\small\itshape\color{black!75}}

\maketitle

\begin{center}
\small\itshape This is a preprint of the accepted manuscript, not the final published version on Image and Vision Computing.
\end{center}

\renewcommand{\thefootnote}{\fnsymbol{footnote}} 
\footnotetext[1]{Corresponding author (O. Rabasseda). Email address: oriol.rabasseda@gmail.com.}
\renewcommand{\thefootnote}{\arabic{footnote}}

\selectlanguage{english}
\begin{abstract}
    Multimodal Large Language Models (MLLMs) have demonstrated strong zero-shot reasoning across diverse domains, but their potential for Human-Object Interaction (HOI) remains unexplored. Unlike traditional HOI approaches that rely on extensive labeled datasets and fixed interaction taxonomies, MLLMs offer flexible reasoning, making them well-suited for dynamic, real-world scenarios. In this work, we present the first systematic study of MLLMs for HOI recognition and detection within a pipeline that combines an external object detector for entity localization with zero-shot multimodal reasoning for interaction understanding. We introduce a benchmark framework for principled zero-shot evaluation, consisting of (i) a unified taxonomy of zero-shot prompting strategies, and (ii) a methodology that transforms token-level logits into interaction scores, enabling fair comparison with specialized HOI models under standard metrics. To support prompt-based evaluation, we additionally construct new question–answer annotations for standard HOI datasets, transforming them into a format suitable for MLLMs. Building on this, we propose two visual prompting strategies, namely Objects as Context and Black Other Objects, that guide MLLMs to target objects while preserving scene context. Experiments on HICO, HICO-DET, and V-COCO demonstrate state-of-the-art zero-shot performance, achieving 53.50 mAP on HICO and 23.69 mAP on HICO-DET. Overall, our work provides a principled benchmarking framework for MLLMs in HOI, opening the door to new models and datasets and enabling flexible open-world interaction understanding. Code and adapted datasets are publicly available\footnote{\url{https://github.com/oriol-rabasseda/mllms-zs-hoi}}.
\end{abstract}


\blfootnote{© 2026. This manuscript version is made available under the CC-BY-NC-ND 4.0 license https://creativecommons.org/licenses/by-nc-nd/4.0/}

\sloppy

\section{Introduction} \label{sec:intro}
\begin{figure}[ht]
    \centering
    \begin{subfigure}[t]{0.44\linewidth}
        \includegraphics[width=\linewidth]{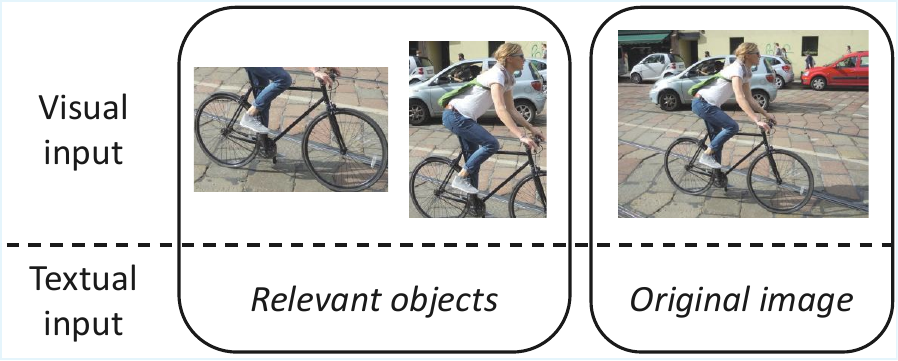}
        \caption{}
        \label{fig:introduction_objects}
    \end{subfigure}
    \begin{subfigure}[t]{0.23\linewidth}
        \includegraphics[width=\linewidth]{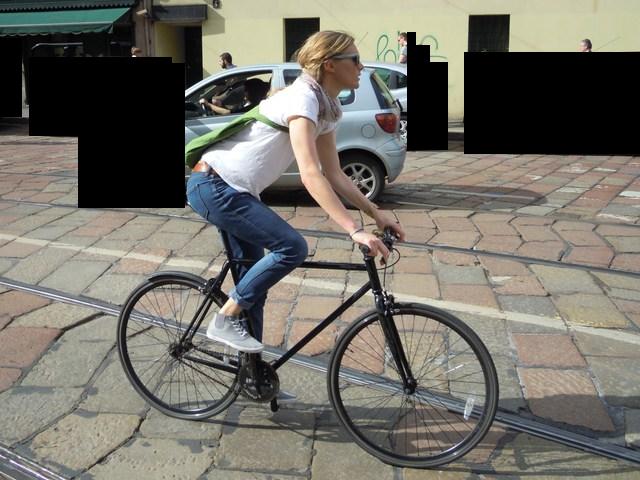}
        \caption{}
        \label{fig:introduction_black}
    \end{subfigure}
    \caption{(a) Visual example of Objects as Context: cropped relevant objects are input alongside text prompts to enrich visual tokens. (b) Visual example of Black Other Objects: all non-target objects are masked to remove distractions.}
\end{figure}

Multimodal Large Language Models (MLLMs) have recently advanced in tasks that connect vision and language, including medical image understanding (\cite{li2023llavamed, moor2023medflamingo}), document parsing (\cite{liu2024textmonkey, hu2024mplugpaperowl}), and embodied agent interaction (\cite{huang2023embodied}). These successes highlight their versatility across diverse benchmarks and domains.

In this work, we investigate MLLMs for Human-Object Interaction (HOI), a core problem in scene understanding. HOI is particularly important for downstream real-world applications such as surveillance, human behavior analysis, and video understanding.

Traditional HOI methods (\cite{yuan2023rlipv2, yang2023boosting, kim2023relational}) rely on large labeled datasets and pre-defined sets of interaction classes. However, such closed-world assumptions limit their ability to generalize: real-world scenarios often contain unseen interactions, novel objects, or evolving vocabularies. For HOI to be broadly useful in practice models must adapt beyond fixed taxonomies.

Prior work on HOI generalization has attempted to address this challenge by withholding certain classes during training and evaluating transfer performance (\cite{shen2018scaling}). However, these approaches remain constrained by labeled training sets and by evaluation protocols that only partially capture open-world conditions. In contrast, MLLMs can operate without task-specific training: by combining multimodal reasoning with natural-language instructions, they enable flexible and interactive zero-shot recognition (\cite{zhang2024multimodal}).

Despite the promise of MLLMs for open-world HOI, a major obstacle is the absence of benchmarks that can fairly evaluate their capabilities. To close this gap, we design the first evaluation framework for MLLMs tailored to HOI recognition and detection in a zero-shot setting. These benchmarks enable systematic comparison with prior HOI methods and provide a foundation for studying MLLMs in realistic open-world conditions.

A key challenge in applying MLLMs to HOI is how to formulate prompts and measure their predictions. To address the prompt design issue, we propose a unified taxonomy of zero-shot prompting strategies for classification and adapt public datasets into a question-answering format suitable for MLLMs. To measure the predictions, we design a method that converts token-level logits into classification scores, allowing direct use of standard metrics such as mAP and ensuring rigorous and fair comparison with specialized HOI models.

Using this framework, we study a broad set of zero-shot prompting methods under our taxonomy and evaluate several lightweight MLLMs. In addition, we introduce two visual prompting methods, namely Objects as Context (Figure \ref{fig:introduction_objects}) and Black Other Objects (Figure \ref{fig:introduction_black}), that guide attention to specified regions while preserving scene context. Overall, we establish strong zero-shot baselines, opening a path toward flexible open-world interaction understanding.

Our main contributions are:

\begin{enumerate}
	\item[(1)] \textbf{A benchmark framework for evaluating MLLMs in HOI}, including (i) a unified taxonomy of zero-shot prompting strategies for classification, and (ii) a method that derives classification scores from token logits, enabling the \textbf{first direct comparison with specialized HOI models}.
	\item[(2)] The construction of novel question-answering annotations for 3 of standard HOI datasets, enabling MLLMs to perform zero-shot classification with natural-language prompts.
	\item[(3)] A comprehensive benchmark of MLLMs and zero-shot strategies on HOI recognition and detection, obtaining state-of-the-art zero-shot results.
	\item[(4)] Two novel visual prompting methods, namely Objects as Context and Black Other Objects, that improve object specification.
\end{enumerate}

The rest of the paper is organized as follows. Section \ref{sec:related_work} reviews related work on MLLMs, zero-shot strategies, and HOI. Section \ref{sec:taxonomy} presents the taxonomy of zero-shot strategies. Section \ref{sec:methodology} outlines our methodology and zero-shot strategies. Section \ref{sec:experiments} reports experiments. Section \ref{sec:conclusions} highlights findings and future directions.

\section{Related work} \label{sec:related_work}
\subsection{Multimodal Large Language Models}
Multimodal Large Language Models (MLLMs) extend Large Language Models (LLMs) to process multimodal inputs, typically text and images. Unlike earlier Vision–Language Models (VLMs), which commonly use fixed encoders and shallow fusion mechanisms for text–image alignment, MLLMs integrate an LLM backbone into the architecture (\cite{yang2407qwen2, cai2024internlm2}). This enables conversational interaction, instruction following, and interleaved multimodal reasoning (\cite{lin2023vila}). Advances in high-resolution encoding (\cite{dong2024internlm, dehghani2023patch}) and efficient backbones have led to competitive models (\cite{yao2024minicpm, li2024llava, wang2024qwen2}).

These models have shown broad versatility across domains, including document parsing (\cite{liu2024textmonkey, hu2024mplugpaperowl}), medical image interpretation (\cite{li2023llavamed, moor2023medflamingo}), and visual reasoning (\cite{liu2023llava, zhu2023minigpt}). Their extensive generalist pretraining makes them effective in zero-shot and open-world settings.

Despite these strengths, MLLMs have not yet been systematically explored for Human–Object Interaction (HOI), a vision task that requires reasoning about localized entities and interaction vocabularies. Existing attempts with vision–language backbones use fixed matching paradigms and lack instruction-following capabilities (\cite{ning2023hoiclip, mao2023clip4hoi}), leaving open the question of how MLLMs perform in zero-shot HOI recognition and detection.

\subsection{Zero-shot prompting strategies for MLLMs}
Early zero-shot work in vision–language models began with CLIP (\cite{radford2021learning}), which showed that classification could be achieved without task-specific training. However, CLIP’s inability to follow detailed instructions or perform multi-step reasoning limited its applicability to complex tasks. These limitations motivated the use of MLLMs for zero-shot inference, particularly in Visual Question Answering (VQA).

Prompting strategies are central to unlocking zero-shot performance and fairly benchmarking MLLMs. They can be broadly divided into textual and visual approaches. Textual strategies aim to improve reasoning and reliability, for instance by generating diverse candidate prompts or constructing them automatically (\cite{lan2023improving}), enriching them with external knowledge (\cite{cao2024knowledge, lin2022revive}), or adopting chain-of-thought templates for step-by-step reasoning (\cite{zhang2024multimodal}). Complementarily, visual prompting modifies the visual input to guide attention, using techniques such as cropping, masking, or altering background context (\cite{shtedritski2023does}).

Despite their promise, most evaluations of these strategies remain confined to VQA benchmarks (\cite{marino2019ok, lu2022learn, goyal2017making}). Although extensions to classification tasks are scarce, studies such as \cite{gupta2024exploring} have explored action recognition with MLLMs, but these efforts typically lack classification score distributions. Without such outputs, standard metrics like mean Average Precision (mAP) cannot be computed, hindering direct comparison with specialized baselines in HOI.

Finally, while several taxonomies of prompting strategies have been proposed (\cite{chen2023unleashing, wu2024visual}), they primarily target textual reasoning or VQA. No prior work systematically unifies textual and visual prompting for classification settings. Our work addresses this gap by introducing a unified taxonomy and evaluation framework tailored to zero-shot HOI.

\subsection{Zero-shot Human-Object Interaction recognition and detection}
Human–Object Interaction (HOI) tasks study how humans and objects interact in images. While recognition focuses on classifying interactions from a fixed set, detection additionally requires localizing both humans and objects. These tasks are particularly challenging due to the large number of possible interactions and the long-tailed, skewed class distributions found in real-world data.

Early HOI recognition and detection methods have advanced through benchmarks such as HICO (\cite{chao2015hico}) and HICO-DET/V-COCO (\cite{chao2018learning, gupta2015visual}), driving progress in both fully supervised and partially zero-shot settings. Partially zero-shot protocols withhold subsets of HOI classes during training (\cite{shen2018scaling, ma2022relvit}), testing limited generalization to unseen interactions. More recently, CLIP-based approaches (\cite{liao2022gen, mao2023clip4hoi}) leverage vision–language pretraining for transfer to unseen categories, demonstrating the potential of multimodal representations for open-world HOI. However, these models still depend on task-specific training subsets and struggle with complex scene-level reasoning.

HOI detection methods are commonly two-stage or one-stage. Two-stage approaches detect humans and objects first, then classify interactions for candidate pairs (\cite{liu2020consnet, zhang2022exploring}), whereas one-stage methods predict interactions and detect objects jointly in an end-to-end model (\cite{liao2022gen, zou2021end}).

In contrast, our work applies instruction-following MLLMs to HOI in a \emph{pure zero-shot} setting. By removing reliance on labeled HOI training data, our benchmarking setup directly tests models under open-world conditions.

\section{Taxonomy of zero-shot strategies for MLLMs} \label{sec:taxonomy}
\begin{figure}[ht]
    \centering
    \includegraphics[width=0.5\linewidth]{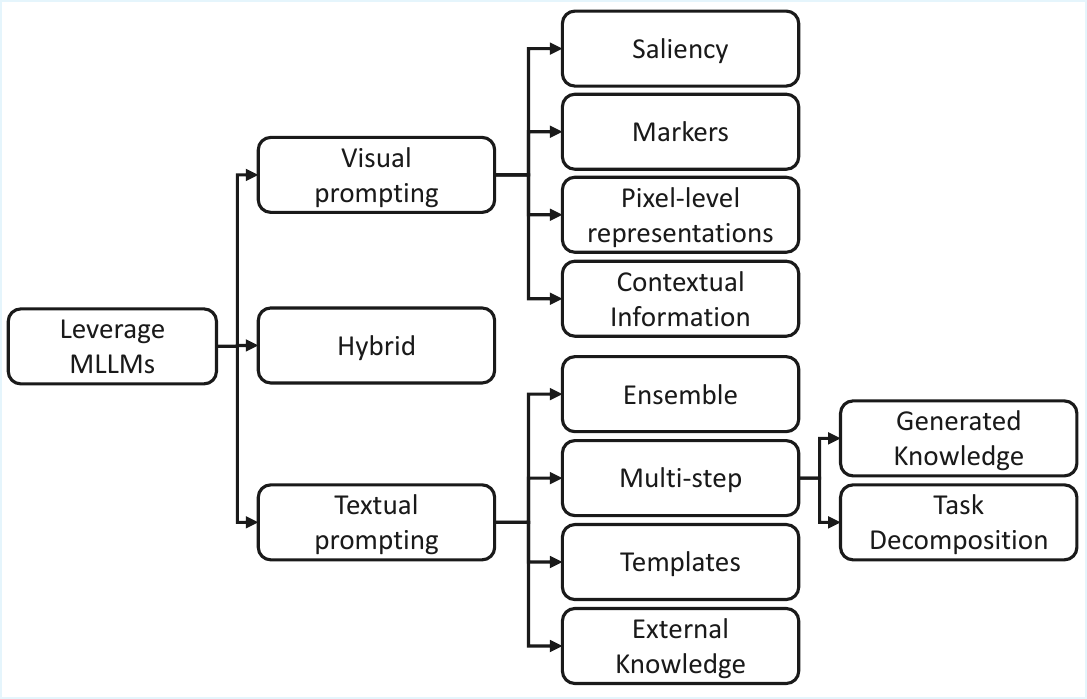}
    \caption{Taxonomy of zero-shot methods leveraging MLLMs divided into textual, visual, and hybrid strategies.}
    \label{fig:taxonomy}
\end{figure}

This section categorizes zero-shot learning strategies to leverage MLLMs as classifiers. We aggregate these techniques based on the type of supplementary information they provide to the model and how it is incorporated into the input of the MLLM. Figure \ref{fig:taxonomy} shows the taxonomy, which is divided into Visual Prompting methods, Textual Prompting methods, and Hybrid methods. Table \ref{tab:taxonomy_methods} categorizes relevant methods under the proposed taxonomy.

\begin{table}[ht]
\centering
\begin{tabular}{@{} p{4cm} p{12cm} @{}}
    \toprule
    Taxonomy & Prompting methods \\ 
    \midrule
    Ensemble & Resampling, Self-consistency (\cite{wang2022self}), Tree-of-thoughts (\cite{yao2023tree}), Graph-of-thoughts (\cite{besta2024graph}) \\
    \midrule
    Generated Knowledge & Multimodal Chain-of-thought (\cite{zhang2024multimodal}), Generated knowledge (\cite{liu2021generated}) \\
    \midrule
    Task Decomposition & Prompt chaining, Least-to-most prompting (\cite{zhou2022leasttomost}) \\
    \midrule
    Templates & Role-prompting, Multimodal Chain-of-thought (\cite{zhang2024multimodal}), Meta prompting (\cite{zhang2023meta}), Prompt Pattern Catalog (\cite{white2023prompt}) \\
    \midrule
    External Knowledge & FLARE (\cite{jiang2023flare}), In-Context RALM (\cite{ram2023context}) \\
    \midrule
    Saliency & RedCircle (\cite{shtedritski2023does}), FGVP (\cite{yang2023fine}) \\
    \midrule
    Markers & RedCircle (\cite{shtedritski2023does}), AutoAD-Zero (\cite{xie2024autoad}), Set-of-Mark (\cite{yang2023set}), OWG (\cite{tziafas2024towards}), Pivot (\cite{nasiriany2024pivot}), VTPrompt (\cite{jiang2024joint}), TextCoT (\cite{luan2024textcot}), CRG (\cite{wan2024contrastive}) \\
    \midrule
    Pixel-level representations & FGVP (\cite{yang2023fine}), SCAFFOLD (\cite{lei2024scaffolding}), AO-Planner (\cite{chen2024affordances}) \\
    \midrule
    Contextual Information & Objects as Context (ours) \\
    \bottomrule
\end{tabular}
\caption{Position in zero-shot taxonomy of relevant methods.}
\label{tab:taxonomy_methods}
\end{table}

\subsection{Textual prompting methods}
Textual prompting focuses on applying specific techniques to the textual part of the multimodal input, so that the MLLM can improve classification by making use of the proper inherent knowledge or the necessary additional information. The prompting methods are categorized into:

\begin{itemize}
    \item \textbf{Templates}: Textual templates consist of reusing specific patterns or sentences that have been deemed useful in similar tasks. Specific templates include role prompting, use of quotes to separate parts of the prompt, and Chain-of-Thought (\cite{kojima2022large}) prompts. Recent methods offer adaptability to multiple tasks, such as pattern catalog (\cite{white2023prompt}) and meta prompting (\cite{zhang2023meta}). Pattern catalog provides an organized collection of prompt templates, while meta prompting focuses on structuring problem formulation.

    \item \textbf{External knowledge}: Concatenating updated or specialized data into the prompt as context is crucial for tasks where the model lacks the required knowledge or information. Providing or retrieving information from a corpus, the Internet or specialized tools fall into this category.
    
    \item \textbf{Ensemble}: Methods that query the LLM in a parallel way belong to this category. Widely used methods include resampling and parallel chain-of-thought-like methods. To fuse the different answers obtained voting strategies and frequency metrics are commonly used.

    \item \textbf{Multi-step}: Multi-step methods query the MLLM several times in a sequential way. We distinguish two strategies: Generated Knowledge and Task Decomposition. Generated Knowledge takes advantage of the knowledge of the MLLM obtained through previous queries and inputs it as context for the final query (\textit{e.g.} image captioning or reasoning from \cite{zhang2024multimodal}). Task Decomposition splits the problem into sub-tasks and merges all the sub-tasks solutions to solve the main task.
\end{itemize}

\subsection{Visual prompting methods}
Visual prompting methods manipulate images and videos to provide additional information, enhancing the visual perception capabilities of the MLLM. Visual prompting strategies are categorized into Markers, Pixel-level Representations, Saliency methods and Contextual Information.

\begin{itemize}    
    \item \textbf{Markers}: Overlay markers such as bounding boxes, circles, arrows, and dots to highlight, identify, or draw attention to particular regions in the visual input. Markers help the MLLM understand the image content and correlate it with the corresponding text.

    \item \textbf{Pixel-level representations}: Overlay fine-grained pixel-level information on the visual input, in opposition to coarse markers or boxes. Object contour highlighting and semantic segmentation masking are examples of this category.
    
    \item \textbf{Saliency}: Modify or delete non-relevant parts of the visual input. Compared to other categories, saliency removes potentially misleading information instead of emphasizing relevant regions. Background adjustments fall into this category (\textit{e.g.} blurring the background).

    \item \textbf{Contextual Information}: Provide the MLLM with additional visual inputs as context, encapsulating relevant information from the original visual input. In this category, the original visual input is not modified; instead, additional information is fed to the MLLM, such as crops of relevant objects.
\end{itemize}

\subsection{Hybrid methods}
Combinations of textual and/or visual prompting methods are possible, as most methods are not mutually exclusive. As an example of a hybrid method, \cite{zhang2024multimodal} proposes Multimodal Chain-of-thought, a two-stage method that uses chain-of-thought templates for the generation of reasoning and inputs it as generated knowledge.

\section{Methodology} \label{sec:methodology}
In this section, we present our methodology for leveraging MLLMs in zero-shot HOI recognition and detection. Section \ref{sec:hoi_recognition} presents our HOI recognition method, including object recognition, MLLM querying, and probability computation. Section \ref{sec:hoi_detection} describes a pipeline that incorporates object detection and object specification strategies. We also introduce two visual prompting methods (Objects as Context and Black Other Objects) designed to guide the model’s attention while maintaining scene context.

\subsection{HOI Recognition} \label{sec:hoi_recognition}
\begin{figure}[ht]
    \centering
    \includegraphics[width=0.9\linewidth]{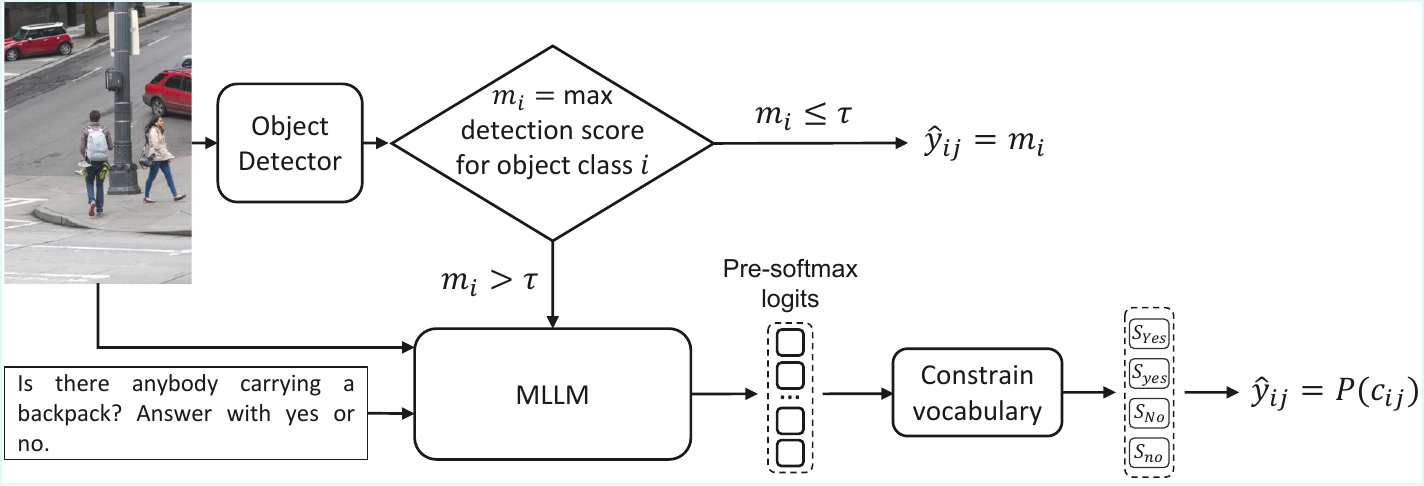}
    \caption{Zero-shot HOI recognition method using MLLMs. The output $\hat{y}_{ij}$ represents the classification score for object class $i$ and interaction $j$, $P(c_{ij})$ denotes the classification probability computed according to Equation \eqref{eq:classification_score}, and $\tau$ is the detection threshold for querying the MLLM. The image is taken from the HICO test set.}
    \label{fig:HICO_architecture}
\end{figure}

Given an input image $x$, the task of HOI Recognition is to predict, for each object class $i \in \mathcal{O}$ (where $\mathcal{O}$ is the set of possible object categories) a probability distribution for each of its potential interactions. Each object $i$ is associated with a predefined set of interactions $\mathcal{I}_i$ (e.g., \textit{holding}, \textit{riding}, or \textit{carrying} depending on the object). To account for cases where an object does not participate in any interaction, this set is sometimes extended with a \textit{no\_interaction} label, yielding $\mathcal{I}_i' = \{no\_interaction\} \cup \mathcal{I}_i$.

For each object $i$, the model produces a vector of predicted probabilities $\hat{y}_i=(\hat{y}_{i1}, \hat{y}_{i2}, ..., \hat{y}_{i|\mathcal{I}_i'|}) \in [0, 1]^{|\mathcal{I}_i'|}$, where $\hat{y}_{ij}$ represents the likelihood that interaction $j \in \mathcal{I}_i'$ occurs with object class $i$. Since multiple interactions can co-occur for the same object and across different objects in the image, HOI Recognition is inherently a multi-label problem. The total output dimensionality is therefore $\sum_{i \in \mathcal{O}} |\mathcal{I}_i'|$. Formally, the objective is to learn a mapping $\textit{f}: \mathcal{X} \to \prod_{i \in \mathcal{O}} [0, 1]^{|\mathcal{I}_i'|}$ from the input space of images $\mathcal{X}$ to the interaction probability vectors, which can be evaluated against ground-truth annotations using standard multi-label metrics.

Our method, including object detection and MLLM-based interaction evaluation, is illustrated in Figure \ref{fig:HICO_architecture}. To improve inference efficiency, an object detector is used to identify which object classes are present in the image and skip unnecessary queries to the MLLM. Specifically, for each detected object class $i$, the maximum detection score $m_i$ among all instances of that class is used as an indicator of presence. If $m_i \leq \tau$, the object class is considered absent and all interactions in $\mathcal{I}_i'$ are assigned this low score ($\hat{y}_{ij} = m_i$), where $\tau$ is a hyperparameter of our method. Equivalently, if an object class $i$ is not detected, $\hat{y}_{ij} = 0$ for all $j \in \mathcal{I}_i'$.

If $m_i > \tau$, the object class $i$ is treated as present and the MLLM is queried to evaluate all interactions in $\mathcal{I}_i'$. Interaction queries are constructed by expressing each interaction in gerund form (e.g., \textit{holding}, \textit{carrying}) and inserting it into the template: ``Is there anybody \{gerund\} a \{object\}? Answer with yes or no.''. Given this instruction, the MLLM is prompted to produce a single-word answer, either affirmative or negative. This formulation yields one independent prediction for each object--interaction pair, allowing multiple interactions to be recognized simultaneously for the same object class.

Although the MLLM produces a textual output, we do not use the actual generated text to compute classification scores. This is because the textual answer only provides a single binary label, which does not produce a probability distribution reflecting the underlying confidence of the model. Instead, we extract the pre-softmax logits of the first generated token, which is sufficient because this token determines whether the answer is affirmative or negative.

We then constrain the vocabulary to a set of affirmative tokens $\mathcal{A}$ and negative tokens $\mathcal{N}$ and compute probabilities over this restricted set using a standard softmax. This effectively treats the restricted logits as a complete vocabulary and allows us to produce classification scores from the likelihood of an affirmative answer versus a negative answer. The specific contents of the sets $\mathcal{A}$ and $\mathcal{N}$ are discussed in Section \ref{sec:class_soundness}.

Formally, the classification score for interaction $c_{ij}$, where $i$ is the object and $j$ the interaction, is computed as:

\begin{equation} \label{eq:classification_score}
    \hat{y}_{ij} = P(c_{ij}) = \frac{\sum_{w \in A} \exp(s_{w})}{\sum_{w \in A} \exp(s_{w}) + \sum_{w \in N} \exp(s_{w})},
\end{equation}

where $s_w$ denotes the logit corresponding to token $w$. Although HOI recognition is a multi-label problem, sometimes object classes include a \textit{no\_interaction} label that should be mutually exclusive with its interactive labels. Estimating this label independently could therefore produce contradictory predictions in which the same object is classified as both interacting and not interacting. To avoid this issue, its probability is computed as the probability that none of the interactions associated with object class $i$ occurs. Consequently, the \textit{no\_interaction} confidence decreases whenever the confidence of one or more interactive labels increases. Formally, where there is a \textit{no\_interaction} class,

\begin{equation} \label{eq:no_interaction}
    P(c_{i,no\_interaction}) = \prod_{j \in \mathcal{I}_i} (1 - P(c_{ij})).
\end{equation}

\begin{figure}[ht]
    \centering
    \includegraphics[width=0.7\linewidth]{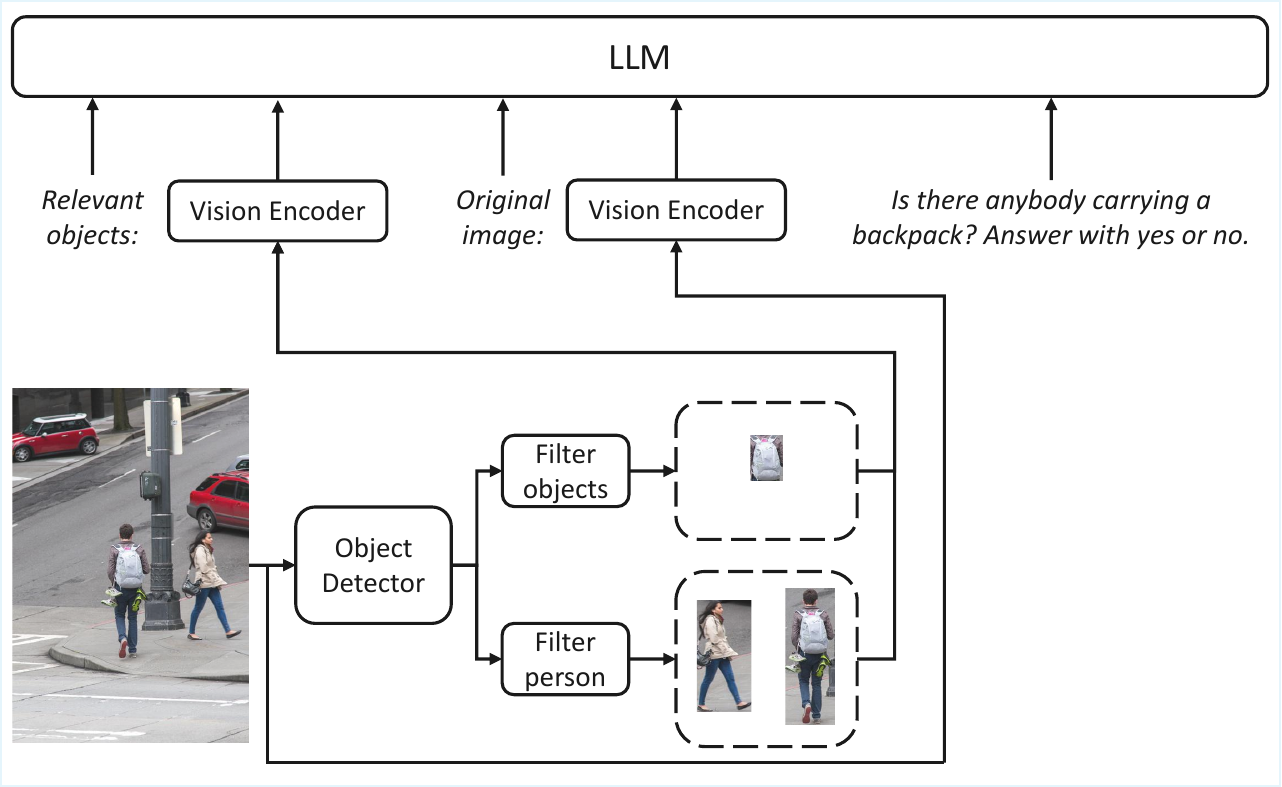}
    \caption{Objects as Context architecture. Step 1: Detect all objects. Step 2a: Filter relevant objects (\textit{e.g.} ``backpacks'') based on the question. Step 2b: Filter all persons. Step 3: Input the information into the LLM, specifying visual tokens related to relevant objects/persons and the original image.}
    \label{fig:objects_diagram}
\end{figure}

Building on this pipeline we evaluate a set of zero-shot prompting strategies drawn from the taxonomy in Section \ref{sec:taxonomy}. The complete list of strategies and their use in recognition is provided in Table \ref{tab:prompting_strategies}, with the corresponding prompts in the Appendix. Multimodal Chain-of-Thought (MCoT) and Caption as Context are two-stage prompting approaches, where an initial prompt generates meaningful information that is incorporated as context for the final binary interaction prediction. On the other hand, Ensemble of Prompts averages the predicted scores obtained from multiple prompt templates. We compare all strategies against a baseline with no additional prompting strategy.

Furthermore, we introduce a visual prompting method, \emph{Objects as Context}, which falls under Contextual Information. In this strategy, the detected regions corresponding to persons and objects of the target class are cropped and supplied as additional visual inputs to the MLLM. This approach aims to focus the model on the entities involved with the queried interaction while still retaining the overall scene context. To avoid overloading the input with less relevant detections, only the $k_1$ most confidently detected target objects and the $k_2$ largest persons (foreground) are included, where $k_1$ and $k_2$ are hyperparameters. Figure \ref{fig:objects_diagram} illustrates the resulting architecture.

\subsection{HOI Detection} \label{sec:hoi_detection}
HOI Detection extends HOI Recognition by requiring not only the classification of which interactions occur, but also the localization of the human–object pairs involved. In recognition, the object detector is used only for filtering and does not constrain the MLLM from handling the task end-to-end. In detection, however, it becomes an integral component of the pipeline, providing bounding boxes for both humans and objects.

Formally, given an input image $x \in \mathcal{X}$, let $\mathcal{H}=\{h_1,...,h_{N_h}\}$ denote the set of detected human bounding boxes and $\mathcal{B}=\{o_1,...,o_{N_o}\}$ the set of detected object bounding boxes provided by the external detector. The task is to predict, for each human–object pair $\langle h_n, o_m \rangle$ and each interaction $j \in \mathcal{I}_{o_m}'$, a probability indicating whether the interaction occurs. Since multiple interactions can co-occur for the same pair (e.g \textit{holding} and \textit{carrying} a \textit{bicycle}), the task is inherently multi-label at the pair level.


Our method follows a two-stage approach: the object detector first provides bounding boxes for humans and objects, after which the MLLM evaluates candidate pairs $\langle h_n, o_m \rangle$. To reduce inference cost and avoid including less relevant detections, we do not evaluate all possible pairs. Instead, we first discard humans and objects whose detection confidence falls below the threshold $\tau$ at instance level. Drawing inspiration from the strategy Objects as Context, we use the tresholds $k_1$ and $k_2$, retaining only the $k_1$ most confidently detected objects and the $k_2$ largest detected persons.

The interaction classification stage mirrors the procedure described for HOI Recognition: queries are expressed in gerund form and appended with the instruction ``Answer with yes or no.'', logits of the first token are extracted, and probabilities are computed over the affirmative ($\mathcal{A}$) and negative ($\mathcal{N}$) token sets. The score for \textit{no\_interaction} is computed following the same formulation as in Equation \ref{eq:no_interaction}, where applicable.

The main challenge is then to specify to the MLLM which human and object are being queried. To address this, we explore different visual and textual prompting strategies for object specification.

Among these strategies, we propose two visual prompting methods. \emph{Objects as Context} is adapted from recognition to only filter the specified object and person. \emph{Black Other Objects} (a Saliency method) masks all detected objects other than the queried human and object, aiming to reduce distractions while preserving the scene background. In addition, we also test established object-specification strategies adapted from \cite{shtedritski2023does}. A complete list of all strategies along with their taxonomy in Section \ref{sec:taxonomy} and their use in detection appears in Table \ref{tab:prompting_strategies}.

\begin{table}[ht]
    \centering
    \begin{tabular}{p{5.7cm} p{5.4cm} c c}
        \toprule
        \textbf{Category (from Taxonomy)} & \textbf{Strategy} & \makecell{\textbf{HOI} \\ \textbf{Recognition}} & \makecell{\textbf{HOI} \\ \textbf{Detection}} \\
        \midrule
        \textbf{Hybrid} & Multimodal Chain-of-Thought (\cite{zhang2024multimodal}) & \checkmark & -- \\
        \textbf{Textual – Generated Knowledge} & Caption as Context & \checkmark & -- \\
        \textbf{Textual – Ensemble} & Ensemble of Prompts & \checkmark & -- \\
        \textbf{Textual – External Knowledge} & Textual Center Coordinates, Textual Boundary Coordinates & -- & \checkmark \\
        \textbf{Visual – Contextual Information} & Objects as Context (ours) & \checkmark & \checkmark \\
        \textbf{Visual – Markers} & Overlay Bounding Boxes, Overlay Elliptic Markers & -- & \checkmark \\
        \textbf{Visual – Saliency} & Blur the Background, Gray the Background, Crop\footnotemark, Black Other Objects (ours) & -- & \checkmark \\
        \bottomrule
    \end{tabular}
    \caption{Summary of prompting strategies evaluated in our benchmark, grouped by the taxonomy in Section \ref{sec:taxonomy}.}
    \label{tab:prompting_strategies}
\end{table}
\footnotetext{The relevant area cropped encompasses both the target object and person.}

\begin{figure}[ht]
    \centering
    \begin{subfigure}[t]{0.145\textwidth}
        \centering
        \includegraphics[width=\linewidth]{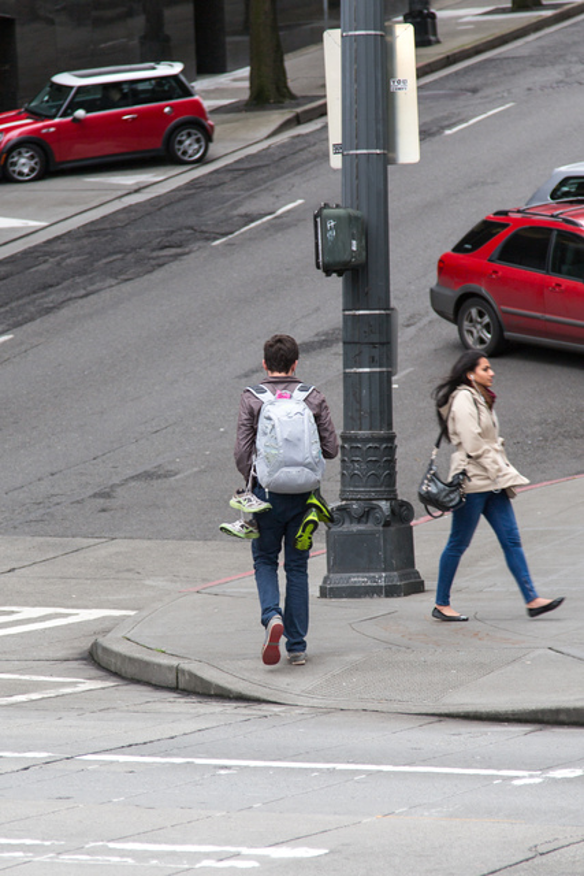}
        \caption{}
    \end{subfigure}
    \begin{subfigure}[t]{0.145\textwidth}
        \centering
        \includegraphics[width=\linewidth]{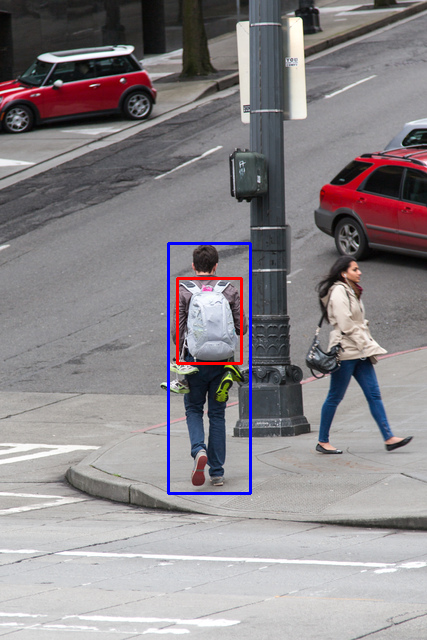}
        \caption{}
    \end{subfigure}
    \begin{subfigure}[t]{0.145\textwidth}
        \centering
        \includegraphics[width=\linewidth]{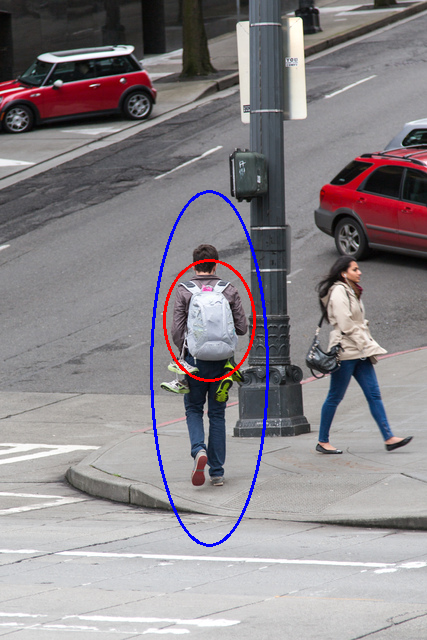}
        \caption{}
    \end{subfigure}
    \begin{subfigure}[t]{0.145\textwidth}
        \centering
        \includegraphics[width=\linewidth]{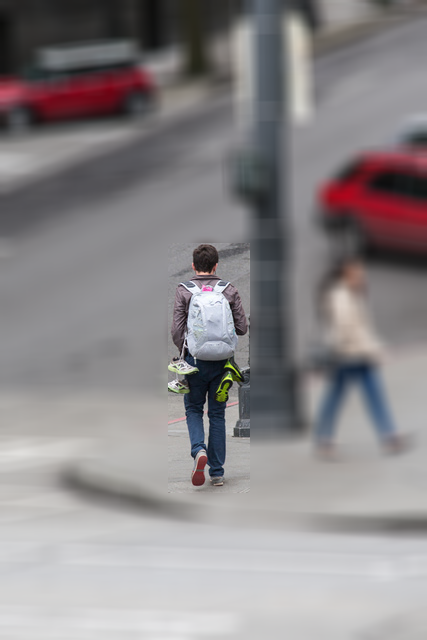}
        \caption{}
    \end{subfigure}
    \begin{subfigure}[t]{0.0712\textwidth}
        \centering
        \includegraphics[width=\linewidth]{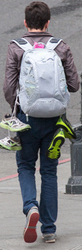}
        \caption{}
    \end{subfigure}
    \begin{subfigure}[t]{0.145\textwidth}
        \centering
        \includegraphics[width=\linewidth]{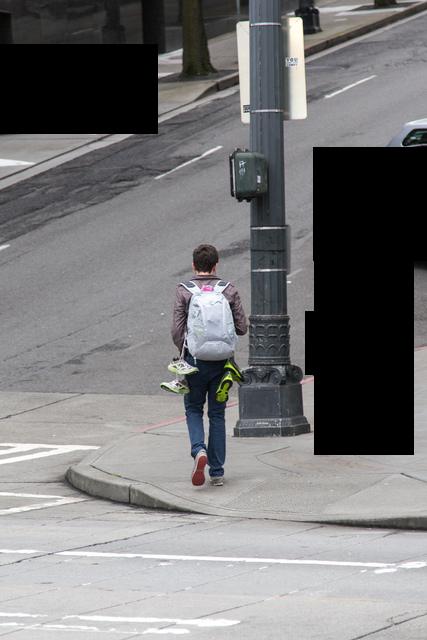}
        \caption{}
    \end{subfigure}
    \begin{subfigure}[t]{0.145\textwidth}
        \centering
        \includegraphics[width=\linewidth]{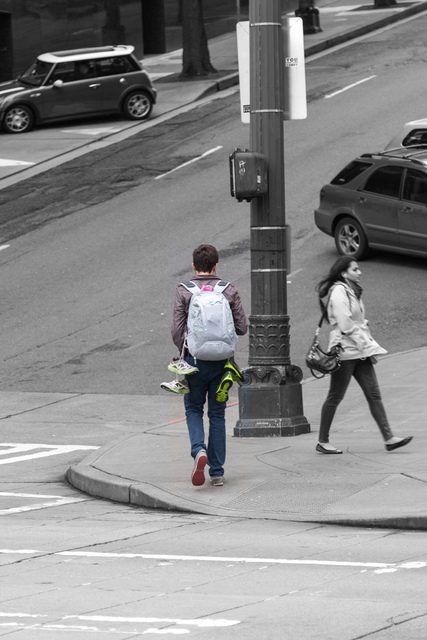}
        \caption{}
    \end{subfigure}
    \caption{Example of all visual prompting methods side-by-side compared to the base image (a). (b) Overlay Bounding Boxes highlights humans and objects with rectangular boxes. (c) Overlay Elliptic Markers uses ellipses to indicate the queried regions. (d) Blur the Background removes contextual clutter by blurring everything outside the target entities. (e) Crop focuses only on the cropped human–object pair. (f) Black Other Objects masks all objects except the queried pair while preserving the background. (g) Gray the Background reduces distraction by desaturating non-target regions.}
    \label{fig:visual_prompt_examples}
\end{figure}

All methods are compared against the baseline, which does not apply any object specification strategy. Examples of textual questions and visual prompts for each strategy are provided in the Appendix and illustrated in Figure \ref{fig:visual_prompt_examples}.

\section{Experiments} \label{sec:experiments}
\subsection{Implementation details}
The object detector used is the extra-large YOLO11 model (\cite{khanam2024yolov11}), pre-trained on the MS-COCO dataset (\cite{lin2014microsoft}).

For HOI recognition, the hyperparameters are set to $\tau=0.1$, $k_1=10$, and $k_2=3$ (see Section \ref{sec:ablation_studies}). For HOI detection, $k_1$ and $k_2$ remain unchanged, while $\tau$ is increased to 0.5 to reduce the number of MLLM queries.

Inference uses greedy decoding with a single output token and is run on three NVIDIA GeForce RTX 3090 GPUs. Additional inference hyperparameters are listed in the Appendix.

\subsection{HOI recognition} \label{sec:hoi_recognition_res}
We construct a HICO Q\&A dataset by adding question–answer annotations for each object–interaction pair, with an illustrative example shown in Figure \ref{fig:HICO_architecture}. The dataset (\cite{chao2015hico}) contains 47,776 images annotated with 600 HOI categories (combinations of 117 action verbs and 80 object categories from MS-COCO), and the original split provides 38,118 training and 9,658 test images. In the \emph{pure zero-shot} evaluation protocol, only the test set is used, with no human–object interactions for fine-tuning. This setting effectively mimics an open-world environment and, to our knowledge, has not previously been applied to the HICO dataset.

Performance is measured by the mean Average Precision (mAP) over HOI classes, as proposed in the original paper, because HOI categories are not mutually exclusive.

\subsubsection{Baseline and score soundness} \label{sec:class_soundness}
We evaluate the lightweight versions of state-of-the-art MLLM families, namely MiniCPM-V (\cite{yao2024minicpm}), LLaVA-OneVision (\cite{li2024llava}), Phi-3.5-Vision (\cite{abdin2024phi}), InternVL2.5 (\cite{chen2023internvl, chen2024far, chen2024expanding}), and Qwen2-VL (\cite{bai2023qwen, wang2024qwen2}). These models vary in how they answer the prompt ``Answer with yes or no'', with differences mainly in capitalization (Yes'' vs. yes'' or No'' vs. no''). To ensure consistency and fair comparison across models, we standardize evaluation by defining the affirmative and negative token sets as $\mathcal{A}=\{\text{``yes''},\text{``Yes''}\}$ and $\mathcal{N}=\{\text{``no''},\text{``No''}\}$.

Although the classification score in Equation \eqref{eq:classification_score} is computed from the logits restricted to $\mathcal{A} \cup \mathcal{N}$, it is important that the tokens in these sets also receive the highest probability under the model’s distribution. This condition ensures that the restricted softmax faithfully reflects the model’s intended prediction. To verify this assumption, and to confirm the robustness gained by including both lowercase and uppercase variants, we evaluate two indicators of score soundness:

\begin{itemize}
    \item \textbf{Different Answers}: frequency of output tokens outside $\mathcal{A} \cup \mathcal{N}$. These tokens are considered to deviate from the expected yes/no answer format.
    \item \textbf{Incoherences}: cases where the token with highest confidence is in $\mathcal{A}$ (i.e. ``Yes'' or ``yes'') but the classification score (from Equation \eqref{eq:classification_score}) is below 0.5, or conversely when the token is in $\mathcal{N}$ (``No'' or ``no'') but the score exceeds 0.5. This indicates a mismatch between the expected model's output and the obtained classification score.
\end{itemize}

\begin{table}[ht]
    \centering
    \begin{tabular}{@{} l c c c @{}}
        \toprule
        Model & mAP $\uparrow$ & Diff. Answers & Incoherences \\
        \midrule
        MiniCPM-V 1 & 48.20 & 0\% & 10.06\% \\
        MiniCPM-V 2 & 47.48 & 0\% & 2.74\% \\
        LLaVA-OV-0.5B (OV) & 46.15 & 0\% & 1.25\% \\
        LLaVA-OV-0.5B (SI) & 44.61 & 0\% & 2.27\% \\
        Phi-3.5-Vision & 40.44 & 7.05\% & 0.00\% \\
        InternVL2.5-1B & 41.81 & 7.44\% & 0.48\% \\
        InternVL2.5-2B & 43.48 & 7.45\% & 0.39\% \\
        Qwen2-VL 2B & 51.39 & 0\% & 0.80\% \\
        Qwen2-VL 7B & \textbf{52.89} & 0\% & 0.48\% \\
        \bottomrule
    \end{tabular}
    \caption{Comparison of MLLMs on the HICO dataset: performance and score soundness study.}
    \label{tab:model_comparison}
\end{table}

Table \ref{tab:model_comparison} shows the mAP values for the baseline computations (without zero-shot strategies from Table \ref{tab:prompting_strategies}) along with the soundness metrics (i.e., Different Answers and Incoherences) on the HICO dataset. The results reveal that, with few exceptions in InternVL2.5 and Phi-3.5-Vision, other tokens never receive the highest output score. This validates the focus on the defined tokens given the prompt instructions. Moreover, the low incidence of incoherences supports the use of both cases (``Yes''/``yes'' and ``No''/``no'') to enhance generalization across models.

The top-performing families are Qwen2-VL and MiniCPM-V. To streamline experiments on zero-shot strategies, we select MiniCPM-V 1 and Qwen2-VL 2B for subsequent tests. Additionally, LLaVA-OV-0.5B (OV), which is trained on multi-image data, is included in experiments that utilize multi-image prompting, namely Objects as Context.

\subsubsection{Zero-shot HOI recognition}
\begin{table}[ht]
    \centering
    \begin{tabular}{@{} l c c c @{}}
        \toprule
        Strategy & MiniCPM-V 1 & Qwen2-VL 2B & LLaVA-OV 0.5B (OV) \\
        \midrule
        MCoT & 39.45 & 30.94 & - \\
        Caption as Context & 44.81 & 49.43 & - \\
        Objects as Context & 44.54 & 53.50 & 47.79 \\
        \bottomrule
    \end{tabular}
    \caption{Evaluations on HICO dataset using zero-shot strategies.}
    \label{tab:zs_hico}
\end{table}

\begin{figure}[ht]
    \centering
    \begin{tikzpicture}
        \begin{axis}[
            width=0.6\textwidth,
            height=7cm,
            xlabel={Number of prompts},
            ylabel={mAP},
            xtick={1, 3, 5, 7, 9},
            ymin=45,
            ymax=55,
            legend pos=south east,
        ]
            \addplot+[mark=*, color=blue, nodes near coords] coordinates {
                (1, 51.39) (3, 53.37) (5, 53.41) (7, 53.29) (9, 53.40)
            };
            \addplot+[mark=square*, color=red, nodes near coords] coordinates {
                (1, 48.20) (3, 50.82) (5, 50.78) (7, 50.27) (9, 49.70)
            };
            \legend{Qwen2-VL 2B, MiniCPM-V 1}
        \end{axis}
    \end{tikzpicture}
    \caption{Evaluation on HICO dataset with different number of prompts using Ensemble of Prompts.}
    \label{fig:ensemble}
\end{figure}
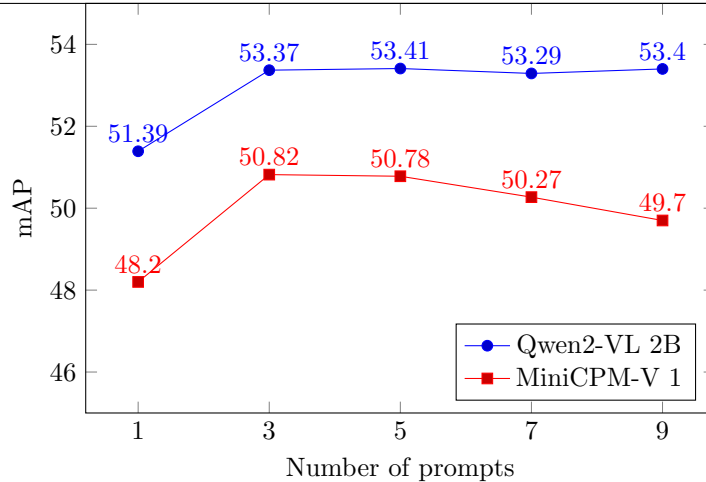

Table \ref{tab:zs_hico} reports the performance of zero-shot strategies on HICO. MCoT and Caption as Context lead to lower performance compared to the baselines in Table \ref{tab:model_comparison}, whereas Objects as Context improves results for models that support multi-image inputs (Qwen2-VL and LLaVA-OneVision). Figure \ref{fig:ensemble} demonstrates that using Ensemble of Prompts mitigates the bias of a single default prompt and improves mAP, though gains plateau beyond three prompts. The lower mAP of the two-stage strategies can be attributed to their tendency to produce more confident predictions conditioned on the generated text. Because the second stage treats the context provided by the first stage as fixed, uncertainty in the initial generation is not propagated to the final score. Consequently, the model may produce low-uncertainty predictions even when the generated context is ambiguous or incorrect, penalizing misclassified samples more severely under the mAP metric. As analyzed in the Appendix, these methods achieve comparable performance after thresholding, indicating reasonable binary decision quality despite less favorable score rankings. When grouped by taxonomy, both Ensemble and Visual Prompting (Objects as Context) outperform the Multi-step with Generated Knowledge methods (MCoT and Caption as Context).

We analyze whether the combination of the two methods that improve performance yields even higher results. However, combining Ensemble of Prompts with Objects as Context (forming a Hybrid approach in our taxonomy) using Qwen2-VL 2B results in a mAP of 51.71, which is lower than when each method is used individually. Therefore, the best \emph{pure zero-shot} result is obtained with Qwen2-VL 2B using Objects as Context, reaching a mAP of \textbf{53.50}. As summarized in Table \ref{tab:comparison_methods}, this result surpasses the strongest partially zero-shot method, which reports 42.0 mAP, highlighting the competitiveness of the proposed \emph{pure zero-shot} evaluation protocol.

\begin{table}[ht]
    \centering
    
    \begin{tabular}{@{} l | c @{}}
    \toprule
    Method & mAP \\
    \midrule
    \textit{Supervised methods} & \\
    R*CNN (\cite{gkioxari2015contextual}) & 28.5 \\
    PaStaNet (\cite{li2020pastanet}) & 46.3 \\
    DEFR (\cite{jin2022overlooked}) & \textbf{65.6} \\
    \midrule
    \textit{Partially zero-shot} & \\
    VCL$^\dagger$ (\cite{hou2020visual}) & 26.7 \\
    RelViT (\cite{ma2022relvit}) & \textbf{42.0} \\
    \midrule
    \textit{Pure Zero-shot} & \\
    \textbf{Qwen2-VL 2B with Objects as Context} & \textbf{53.50} \\
    \bottomrule
    \end{tabular}
    \caption{Comparison with prior work for HOI recognition on HICO. \textit{Note:} $^\dagger$ results are reported in \cite{ma2022relvit}.}
    \label{tab:comparison_methods}
\end{table}

\subsection{HOI detection}
We conduct experiments on two datasets for HOI detection: HICO-DET (\cite{chao2018learning}) and V-COCO (\cite{gupta2015visual}). Based on these datasets, we construct HICO-DET Q\&A and V-COCO Q\&A versions by adding question–answer annotations for each object–interaction pair, enabling evaluation with MLLMs in a prompt-based format (an illustrative example is shown in Figure \ref{fig:objects_diagram}).

HICO-DET extends the HICO dataset with 151,276 annotated human–object pairs across 47,776 images, providing bounding boxes for both humans and objects together with interaction labels. The split contains 38,118 training and 9,658 test images. Evaluation follows the standard mean Average Precision (mAP) metric: a detection is considered correct when both the human and object boxes have an Intersection-over-Union (IoU) above 0.5 with the ground truths. Results are reported for three category sets: Full (all 600 HOI categories), Rare (138 categories with fewer than ten training instances), and Non-Rare (462 categories with at least ten instances). Evaluations are conducted in a \emph{pure zero-shot} setting, where no human–object interactions from HICO-DET are used for fine-tuning. \emph{Pure zero-shot} results on HICO-DET have previously been reported (\cite{yuan2022rlip}), providing a useful reference point for our benchmark.

V-COCO is a subset of MS-COCO containing 10,346 images (2,533 train, 2,867 validation, 4,946 test) annotated with 26 action categories. Each action is associated with one or two object roles (e.g., ``cut'' may involve both knife and food). As in HICO-DET, Average Precision is computed for each action–role pair and the mean is reported. The official evaluation code (\cite{gupta2015visual}) defines two variants for handling missing object-role annotations: (1) \({AP}_{role}^{\#1}\), where the object is considered occluded and a dummy bounding box must be predicted, and (2) \({AP}_{role}^{\#2}\), where the object role is treated as out of scope and no bounding box is required. In our pipeline, the MLLM is evaluated without using any V-COCO supervision, while the object detector is pre-trained on MS-COCO. As a result, the \emph{pure zero-shot} setting strictly applies to the MLLM, while the overall pipeline benefits from an object detector trained on data from the same distribution.

To enable MLLM evaluation, we generate a question–answer version of each dataset. For HICO-DET, we create a natural language question for each human–object pair and interaction, formatted to produce a yes/no answer. Because the pipeline requires a predefined set of possible HOI categories with specific object classes and interactions to prompt the MLLM and V-COCO defines only action–role pairs, we construct the set of HOI categories from the action–object combinations found in the \textit{trainval} split. This adaptation does not alter the official evaluation metrics and ensures fair comparison with other methods.

\begin{table}[ht]
    \centering
    \begin{tabular}{@{} l | c c c | c c @{}}
        \toprule
        \multirow{2}{*}{Method} & \multicolumn{3}{c|}{HICO-DET} & \multicolumn{2}{c}{V-COCO} \\
         & Full & Rare & Non-rare & ${AP}_{role}^{\#1}$ & ${AP}_{role}^{\#2}$\\
        \midrule
        \textit{MiniCPM-V 1} & & & & & \\
        Baseline & 20.78 & 26.63 & 19.03 & 19.95 & 21.98 \\
        Objects as Context & 22.17 & 27.53 & 20.57 & 21.46 & 23.09 \\
        Overlay Bounding Boxes & 15.90 & 19.55 & 14.81 & 5.65 & 6.11 \\
        Overlay Elliptic Marker & 17.70 & 22.55 & 16.25 & 5.94 & 6.24 \\
        Blur Background & \textbf{22.46} & 27.83 & \textbf{20.85} & 24.82 & 26.86 \\
        Crop & 20.96 & 26.49 & 19.31 & 23.70 & 26.16 \\
        Black Other Objects & 22.43 & \textbf{28.60} & 20.59 & \textbf{26.16} & \textbf{28.56} \\
        Gray Background & 19.59 & 26.02 & 17.67 & 7.18 & 7.73 \\
        Textual Center Coords. & 13.84 & 14.98 & 13.50 & 6.29 & 6.64 \\
        Textual Boundary Coords. & 15.41 & 20.35 & 13.94 & 13.38 & 15.16 \\
        \midrule
        \textit{Qwen2-VL 2B} & & & & & \\
        Baseline & 20.00 & 27.59 & 17.73 & 17.28 & 19.19 \\
        Objects as Context & 20.56 & 27.50 & 18.48 & 18.03 & 19.95 \\
        Overlay Bounding Boxes & 19.41 & 24.86 & 17.78 & 6.70 & 7.14 \\
        Overlay Elliptic Marker & 19.86 & 25.70 & 18.12 & 5.31 & 5.53 \\
        Blur Background & \textbf{23.69} & \textbf{30.04} & \textbf{21.79} & \textbf{32.27} & \textbf{34.93} \\
        Crop & 18.46 & 25.56 & 16.34 & 19.19 & 21.81 \\
        Black Other Objects & 23.01 & 29.87 & 20.97 & 22.89 & 25.17 \\
        Gray Background & 19.65 & 24.75 & 18.13 & 11.05 & 11.79 \\
        Textual Center Coords. & 12.84 & 14.74 & 12.27 & 4.95 & 6.03 \\
        Textual Boundary Coords. & 15.76 & 20.40 & 14.37 & 18.07 & 20.60 \\
        \midrule
        \textit{LLaVA-OV-0.5B (OV)} & & & & & \\
        Baseline & 18.52 & \textbf{26.82} & 16.04 & 
        16.31 & 18.61 \\
        Objects as Context & \textbf{19.23} & 25.10 & \textbf{17.47} & \textbf{17.65} & \textbf{19.28} \\
        \bottomrule
    \end{tabular}
    \caption{\emph{Pure zero-shot} evaluation on HICO-DET and V-COCO of the prompting strategies for object specification.}
    \label{tab:hoi_approaches_res}
\end{table}

Table \ref{tab:hoi_approaches_res} summarizes the performance of the prompting strategies listed in Table \ref{tab:prompting_strategies}. Across both HICO-DET and V-COCO the trends are consistent: (1) passing object detections as text (Textual Center Coordinates and Boundary Coordinates) or overlaying bounding boxes/elliptic markers degrades performance; (2) saliency-based modifications have mixed effects: Gray the Background and Crop are neutral or negative, whereas Blur the Background and Black Other Objects consistently improve results; and (3) adding Objects as Context provides a modest boost. Overall, these three strategies are the only ones that show positive impact across all models, with Qwen2-VL 2B + Blur the Background emerging as the best configuration on both datasets.

Although the qualitative trends are similar across datasets, the magnitude of the improvements is substantially larger on V-COCO. We attribute this difference to the higher interaction density of V-COCO: on average, its images contain 6.04 detected objects and 4.46 interactions, compared with 5.25 detected objects and 2.10 interactions in HICO-DET. Isolating the target human--object region may be particularly helpful in scenes containing many simultaneous interactions, as it reduces ambiguity among multiple candidate interactions. Apart from the larger magnitude of the changes in V-COCO, some methods also behave slightly differently across the two datasets. Crop improves over the corresponding baseline on V-COCO, whereas it is neutral or detrimental on HICO-DET.  Black Other Objects consistently improves performance by removing potentially distracting visual information. However, in interaction-dense scenes, masking several objects may introduce sharp artificial boundaries. One possible interpretation is that the vision encoders of different MLLMs respond differently to these artifacts. Although our experiments do not isolate this factor, it could explain the model-dependent performance differences observed on V-COCO for Black Other Objects.

Table \ref{tab:hico_det_benchmark} contextualizes the best configuration among representative HOI detection methods under different settings according to its supervision level. Results show that Qwen2-VL 2B with Blur the Background achieves 23.69 mAP on HICO-DET and 32.3/34.9 ${AP}_{role}^{\#1}$/${AP}_{role}^{\#2}$ on V-COCO, surpassing the previously reported \emph{pure zero-shot} method RLIPv2.

The \emph{pure zero-shot} setting is particularly promising for rare categories. Since no model is trained on the evaluation data, the interaction reasoning module does not directly inherit the long-tail class distributions of the benchmark annotations. Consistent with this expectation, Table \ref{tab:hico_det_benchmark} shows that the best configuration performs better on Rare categories than on Non-rare categories on HICO-DET (30.04 vs. 21.79 mAP). Moreover, on the Rare split, it obtains a strong performance compared to the methods leveraging a pre-trained object detector.

\begin{table*}
    \centering
    \begin{tabular}{@{} l | c c c | c c @{}}
        \toprule
        \multirow{2}{*}{Method} & \multicolumn{3}{c|}{HICO-DET} & \multicolumn{2}{c}{V-COCO} \\
        & Full & Rare & Non-rare & ${AP}_{role}^{\#1}$ & ${AP}_{role}^{\#2}$ \\
        \midrule
        \textit{Fully supervised} & & & & & \\
        GEN-VLKT (\cite{liao2022gen}) & 34.95 & 31.18 & 36.08 & 63.6 & 65.9 \\
        SICHOI (\cite{luo2024discovering}) & 45.04 & 45.61 & 44.88 & \textbf{71.1} & \textbf{75.6} \\
        Pose-Aware (\cite{wu2024exploring}) & \textbf{46.01} & \textbf{46.74} & \textbf{45.80} & 63.0 & 68.7 \\
        \midrule
        \textit{Pre-trained object detector} & & & & & \\
        ConsNet (\cite{liu2020consnet}) & 22.15 & 17.55 & 23.52 & - & - \\
        STIP (\cite{zhang2022exploring}) & \textbf{28.81} & \textbf{27.55} & \textbf{29.18} & - & - \\
        \midrule
        \textit{Partially zero-shot} & & & & & \\
        CLIP4HOI (\cite{mao2023clip4hoi}) & 32.11 & - & - & - & - \\
        KI2HOI (\cite{xue2024towards}) & \textbf{34.56} & - & - & - & - \\
        \midrule
        \textit{Pure Zero-shot} & & & & & \\
        RLIPv2 (\cite{yuan2023rlipv2}) & 23.29 & 27.97 & \textbf{21.90} & - & - \\
        \textbf{Qwen2-VL 2B with Blur the background} & \textbf{23.69} & \textbf{30.04} & 21.79 & \textbf{32.3} & \textbf{34.9} \\
        \bottomrule
    \end{tabular}
    \caption{Comparison with other methods on HICO-DET and V-COCO. Fully supervised methods are trained end-to-end with HOI annotations, while methods with a pre-trained object detector reuse a pre-trained object detector and train only the interaction head. Partially zero-shot methods hold out 120 action--object pairs during training and report HICO-DET performance over the full set of 600 classes, without separate Rare and Non-rare results. V-COCO does not define an analogous held-out setting. \emph{Pure zero-shot} methods do not use any HOI training data. Not all methods are evaluated on both datasets; - denotes results that are not reported. We report the best overall configuration from Table~\ref{tab:hoi_approaches_res} as our method.}
    \label{tab:hico_det_benchmark}
\end{table*}

\subsection{HOI detection with ground truth objects} \label{sec:hoi_detection_gt}
To further isolate the effect of the benchmarked prompting strategies, we introduce a \emph{pure zero-shot} evaluation protocol that leverages ground-truth bounding boxes on HICO-DET. This protocol disentangles the contribution of object detection errors from interaction classification performance, allowing a more direct assessment of object specification strategies. While this evaluation setup has previously been applied to supervised HOI methods (\cite{gao2018ican, zhang2022exploring}), our work is the first to explore it in a \emph{pure zero-shot} regime. In addition to reporting the quantitative results, we provide a qualitative analysis of the proposed visual prompting methods to better understand their behavior.

\begin{table}[ht]
    \centering
    \begin{tabular}{@{} l ccc @{}}
        \toprule
        \multirow{2}{*}{\textbf{Method}} & \multicolumn{3}{c}{mAP} \\
         & Full & Rare & Non-rare \\
        \midrule
        \textit{MiniCPM-V 1} & & & \\
        Baseline & 46.35 & 47.79 & 45.32 \\
        Objects as Context & \textbf{48.41} & 50.85 & \textbf{47.68} \\
        Overlay Bounding Boxes & 39.31 & 42.06 & 38.49 \\
        Overlay Elliptic Marker & 42.89 & 45.38 & 42.14 \\
        Blur Background & 45.93 & 47.94 & 45.33 \\
        Crop & \textbf{48.41} & 51.80 & 47.39 \\
        Black Other Objects & 47.64 & \textbf{52.43} & 46.20 \\
        Gray Background & 40.58 & 46.19 & 38.91 \\
        Textual Center Coordinates & 36.80 & 31.57 & 38.37 \\
        Textual Boundary Coordinates & 41.50 & 42.25 & 41.27 \\
        \midrule
        \textit{Qwen2-VL 2B} & & & \\
        Baseline & 49.20 & 53.51 & 47.92 \\
        Objects as Context & \textbf{51.41} & \textbf{56.57} & \textbf{49.86} \\
        Overlay Bounding Boxes & 45.18 & 48.49 & 44.19 \\
        Overlay Elliptic Marker & 45.55 & 47.00 & 45.12 \\
        Blur Background & 46.43 & 51.51 & 44.91 \\
        Crop & 48.83 & 53.86 & 47.34 \\
        Black Other Objects & 50.01 & 55.14 & 48.48 \\
        Gray Background & 42.65 & 47.52 & 41.20 \\
        Textual Center Coordinates & 31.71 & 28.75 & 32.60 \\
        Textual Boundary Coordinates & 39.03 & 42.25 & 38.07 \\
        \midrule
        \textit{LLaVA-OV-0.5B (OV)} & & & \\
        Baseline & 44.06 & 49.15 & 42.53 \\
        Objects as Context & \textbf{47.33} & \textbf{50.67} & \textbf{46.33} \\
        \bottomrule
    \end{tabular}
    \caption{\emph{Pure zero-shot} evaluation on HICO-DET removing the effect of the object detector.}
    \label{tab:hico_det_nood}
\end{table}

Table \ref{tab:hico_det_nood} summarizes the results. Among the evaluated strategies, only Objects as Context and Black Other Objects consistently improve HOI classification across all models relative to their baselines. Overall, using ground-truth bounding boxes yields substantially higher mAP compared to Table \ref{tab:hoi_approaches_res}, confirming that a significant portion of the performance gap arises from detector errors. The best-performing configuration is Objects as Context with Qwen2-VL 2B.

\paragraph{Qualitative analysis:} the following observations help explain why the proposed visual prompting methods (Objects as Context and Black Other Objects) obtain superior results in this setting on HICO-DET. All qualitative analyses are performed using Qwen2-VL 2B, as it is the MLLM reporting better results.

\begin{figure*}
    \centering
    \begin{subfigure}[t]{0.19\linewidth}
        \centering
        \includegraphics[width=\linewidth]{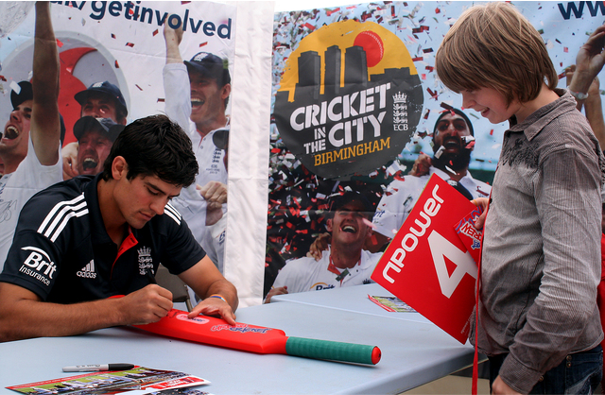}
        \caption{\textit{Sign baseball bat}}
        \label{fig:rank1}
    \end{subfigure}
    \begin{subfigure}[t]{0.19\linewidth}
        \includegraphics[width=\linewidth]{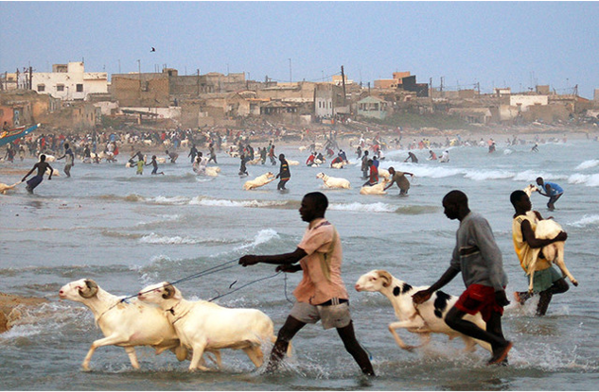}
        \caption{\textit{Wash sheep}}
        \label{fig:rank2}
    \end{subfigure}
    \begin{subfigure}[t]{0.19\linewidth}
        \includegraphics[width=\linewidth]{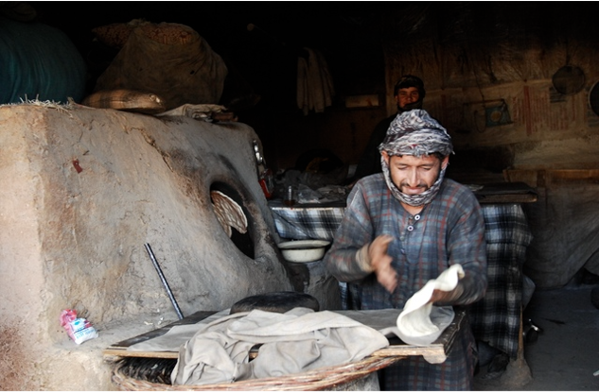}
        \caption{\textit{Repair oven}}
        \label{fig:rank3}
    \end{subfigure}
    \begin{subfigure}[t]{0.19\linewidth}
        \includegraphics[width=\linewidth]{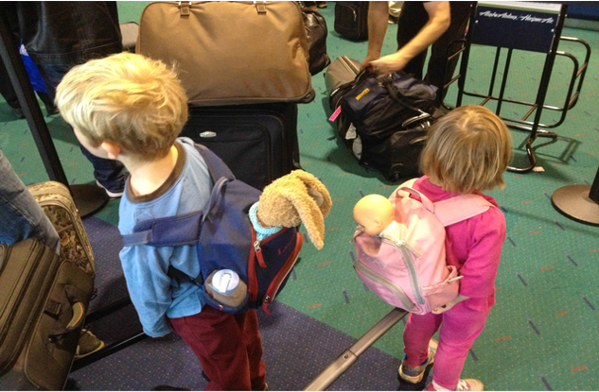}
        \caption{\textit{Inspect backpack}}
        \label{fig:rank4}
    \end{subfigure}
    \begin{subfigure}[t]{0.19\linewidth}
        \includegraphics[width=\linewidth]{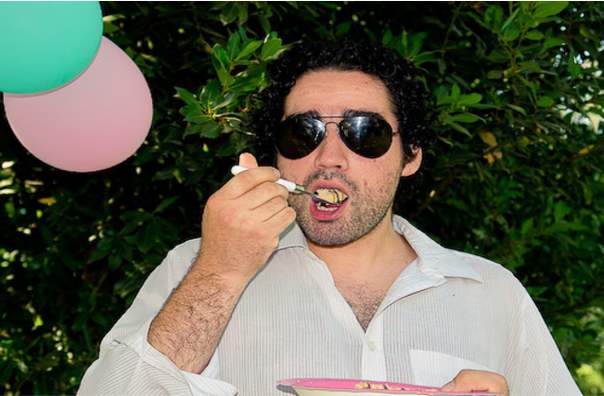}
        \caption{\textit{Eat cake}}
        \label{fig:rank5}
    \end{subfigure}
    \caption{Image examples of the HOIs that exhibit the highest performance gains when applying the identified prompting methods. Objects as Context improves detection for interactions \textit{Sign baseball bat} (+47.69 AP), \textit{Wash sheep} (+29.44 AP) and \textit{Repair oven} (+27.26 AP), while Black Other Objects enhances detection for \textit{Inspect backpack} (+34.43 AP), \textit{Sign baseball bat} (+28.83 AP) and \textit{Eat cake} (+28.75 AP).}
    \label{fig:ranking}
\end{figure*}

\begin{figure*}
    \centering
    \begin{subfigure}[t]{0.165\linewidth}
        \centering
        \includegraphics[width=\linewidth]{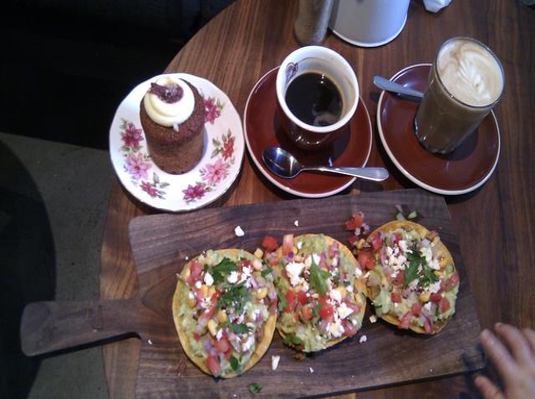}
        \caption{\textit{No\_interaction spoon}}
        \label{fig:rank1_neg}
    \end{subfigure}
    \hspace{-5pt}
    \begin{subfigure}[t]{0.165\linewidth}
        \includegraphics[width=\linewidth]{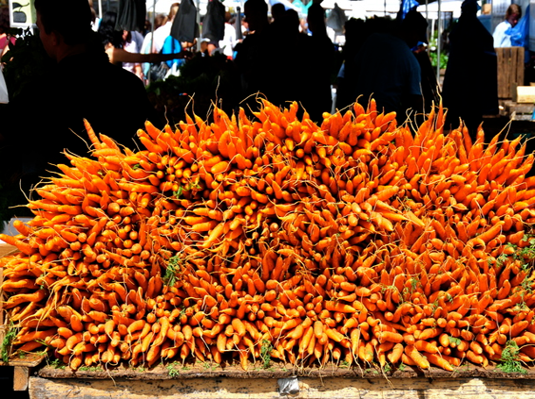}
        \caption{\textit{No\_interaction carrot}}
        \label{fig:rank2_neg}
    \end{subfigure}
    \hspace{-5pt}
    \begin{subfigure}[t]{0.165\linewidth}
        \includegraphics[width=\linewidth]{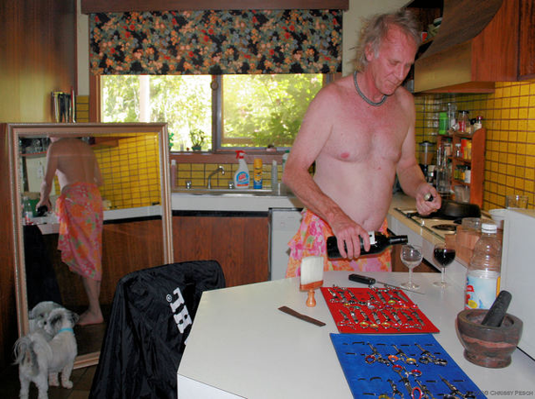}
        \caption{\textit{Fill wine glass}}
        \label{fig:rank3_neg}
    \end{subfigure}
    \hspace{-5pt}
    \begin{subfigure}[t]{0.165\linewidth}
        \includegraphics[width=\linewidth]{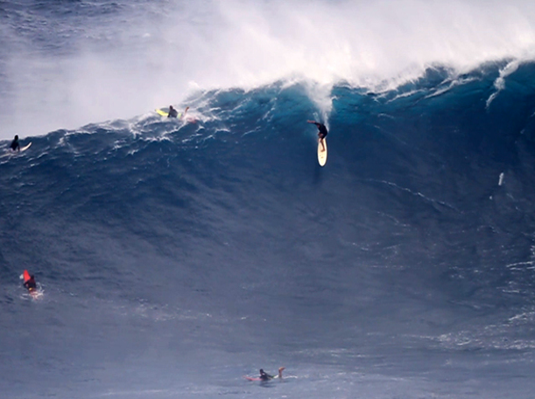}
        \caption{\textit{Lie on surfboard}}
        \label{fig:rank4_neg}
    \end{subfigure}
    \hspace{-5pt}
    \begin{subfigure}[t]{0.165\linewidth}
        \includegraphics[width=\linewidth]{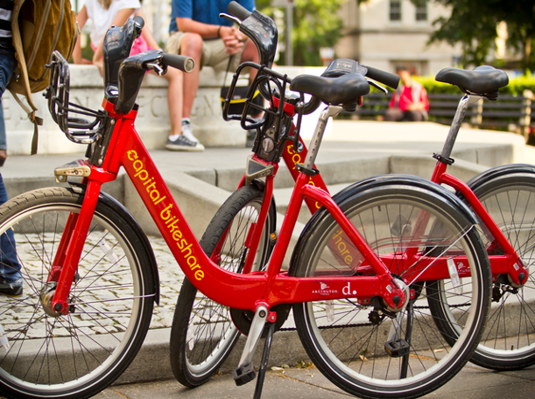}
        \caption{\textit{No\_interaction bicycle}}
        \label{fig:rank5_neg}
    \end{subfigure}
    \hspace{-5pt}
    \begin{subfigure}[t]{0.165\linewidth}
        \includegraphics[width=\linewidth]{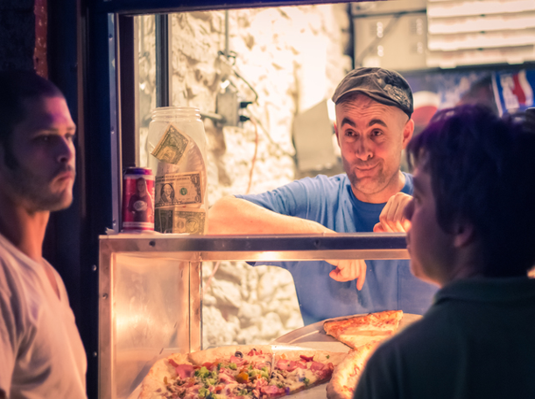}
        \caption{\textit{Buy pizza}}
        \label{fig:rank6_neg}
    \end{subfigure}
    \caption{Image examples of the HOIs that exhibit the highest performance losses when applying the identified prompting methods. Objects as Context suffers detection decrease for interactions \textit{No\_interaction spoon} (-28.90 AP), \textit{No\_interaction carrot} (-21.41 AP) and \textit{Fill wine glass} (-17.62 AP), while Black Other Objects downgrades detection for \textit{Lie on surfboard} (-48.34 AP), \textit{No\_interaction bicycle} (-36.23 AP) and \textit{Buy pizza} (-35.97 AP).}
    \label{fig:ranking_neg}
\end{figure*}

\begin{figure}[ht]
    \centering
    \begin{subfigure}[t]{0.32\linewidth}
        \includegraphics[width=\linewidth]{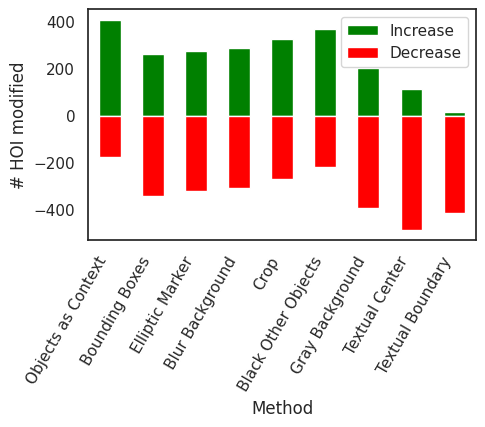}
        \caption{}
        \label{fig:HOI_difference}
    \end{subfigure}
    \begin{subfigure}[t]{0.32\linewidth}
        \includegraphics[width=\linewidth]{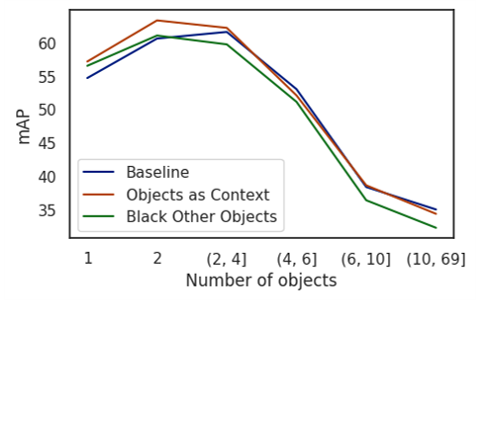}
        \caption{}
        \label{fig:obj_count}
    \end{subfigure}
    \begin{subfigure}[t]{0.32\linewidth}
        \includegraphics[width=\linewidth]{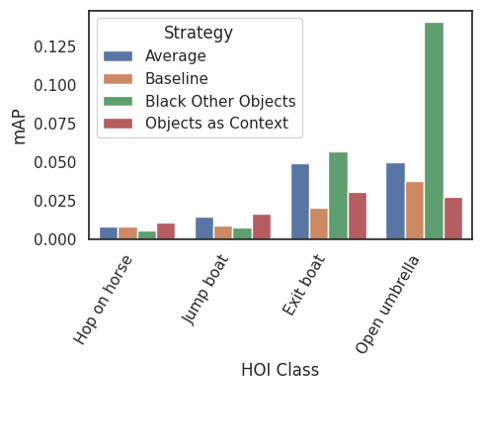}
        \caption{}
        \label{fig:worst_classes}
    \end{subfigure}
    \caption{Impact of prompting methods on HICO-DET using ground truth bounding boxes. (a) Number of HOIs showing an increase or decrease in AP compared to the baseline for all prompting methods. (b) Performance of the proposed methods relative to the number of detected objects of the target class. (c) Performance gains of the proposed methods on the most challenging classes. Most challenging classes are considered those that have a lower average AP over all object specification approaches.}
\end{figure}

\begin{figure}[ht]
    \centering
    \begin{subfigure}[t]{0.32\linewidth}
        \includegraphics[width=\linewidth]{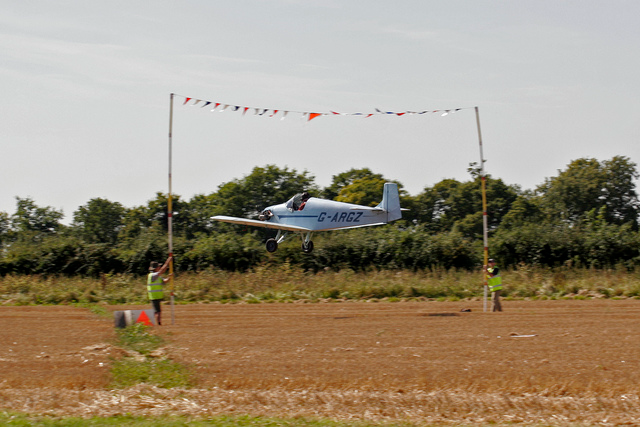}
        \caption{\textit{Sit on plane}}
        \label{fig:use1}
    \end{subfigure}
    \begin{subfigure}[t]{0.16\linewidth}
        \centering
        \includegraphics[width=\linewidth]{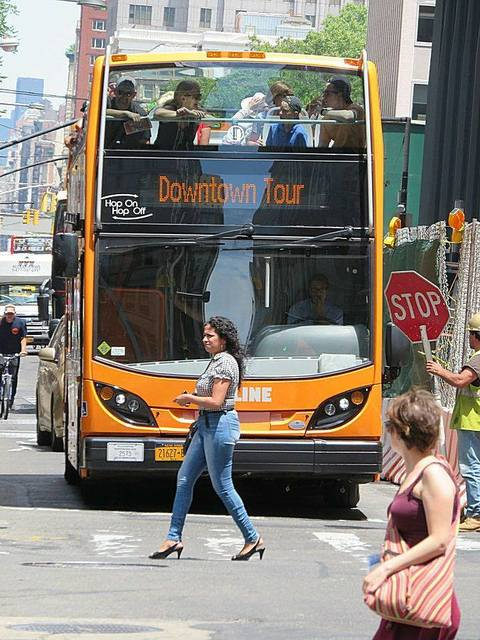}
        \caption{\textit{Direct bus}}
        \label{fig:use2}
    \end{subfigure}
    \begin{subfigure}[t]{0.16\linewidth}
        \centering
        \includegraphics[width=\linewidth]{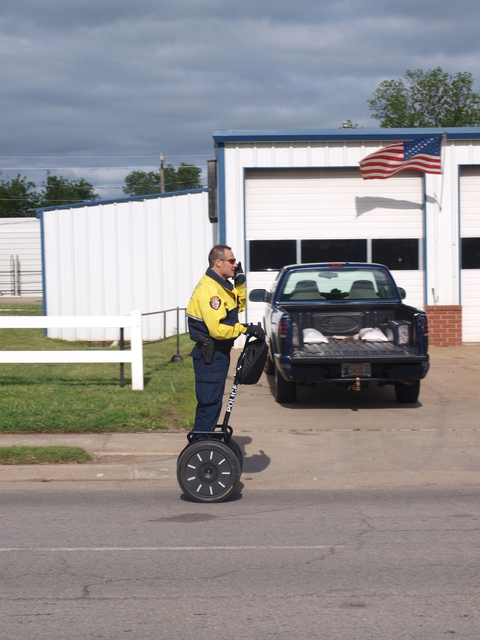}
        \caption{\textit{Carry phone}}
        \label{fig:use3}
    \end{subfigure}
    \begin{subfigure}[t]{0.32\linewidth}
        \includegraphics[width=\linewidth]{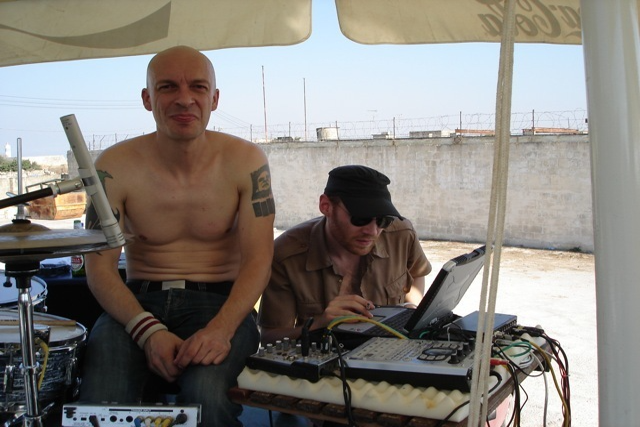}
        \caption{\textit{Read laptop}}
        \label{fig:use4}
    \end{subfigure}
    \par\bigskip
    \begin{subfigure}[t]{0.49\linewidth}
        \begin{subfigure}[t]{0.50\linewidth}
            \includegraphics[width=\linewidth]{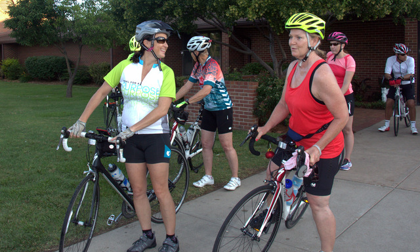}
        \end{subfigure}
        \hspace{-6pt}
        \begin{subfigure}[t]{0.50\linewidth}
            \includegraphics[width=\linewidth]{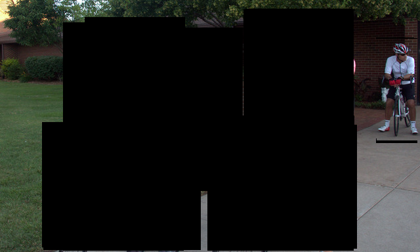}
        \end{subfigure}
        \caption{\textit{Sit on bicycle}}
        \label{fig:use5}
    \end{subfigure}
    \begin{subfigure}[t]{0.49\linewidth}
        \begin{subfigure}[t]{0.50\linewidth}
            \includegraphics[width=\linewidth]{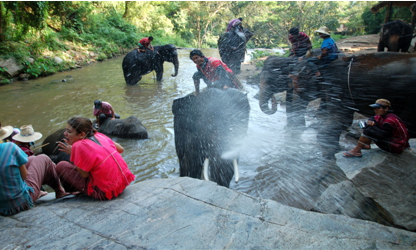}
        \end{subfigure}
        \hspace{-6pt}
        \begin{subfigure}[t]{0.50\linewidth}
            \includegraphics[width=\linewidth]{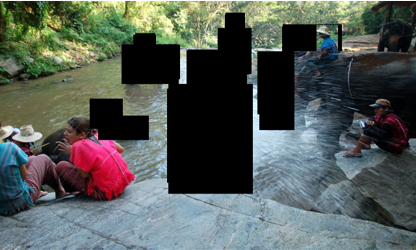}
        \end{subfigure}
        \caption{\textit{Ride elephant}}
        \label{fig:use6}
    \end{subfigure}
    \caption{Examples where Objects as Context (a-d) and Black Other Objects (e-f) correctly classify a HOI that all other methods fail to detect: (a) a focused crop of the pilot, (b) a crop of the person holding the STOP sign, (c) a crop of the small and partially occluded cell phone, and (d) a crop of the laptop and the interacting person presumably help the model with more specific visual context. (e) Masking other cyclists and (f) masking irrelevant overlapping objects presumably direct model’s attention to the correct interaction.}
    \label{fig:use_cases}
\end{figure}

\begin{enumerate}
    \item[\textit{(i)}] Figure \ref{fig:ranking} shows that some HOI classes experience a steep increase in performance. Most of these classes (\textit{Sign baseball bat, Wash Sheep}, and \textit{Inspect backpack}) are within the set of Rare interactions and have less than or equal to five ground truth images in the test set. The increase might be due to the benefits of properly specifying small or background objects (\textit{e.g.} in Figures \ref{fig:ranking}\subref{fig:rank2},\subref{fig:rank4} the action of the HOI class mentioned occurs in the background) or occluded objects (\textit{e.g.} the cake in Figure \ref{fig:ranking}\subref{fig:rank5} is out of the image frame).

    \item[\textit{(ii)}] Figures \ref{fig:ranking_neg}\subref{fig:rank1_neg},\subref{fig:rank2_neg},\subref{fig:rank3_neg} show examples of the HOI classes that experience a larger performance drop when applying Objects as Context for object specification. We identify that the performance decline may occur when either there is no interaction (Figures \ref{fig:ranking_neg}\subref{fig:rank1_neg},\subref{fig:rank2_neg}) or cropping the object deletes crucial information (Figure \ref{fig:ranking_neg}\subref{fig:rank3_neg}). In the former case, crops of the objects lead to higher classification scores of interactive HOIs which, by the construction of the \textit{no\_interaction} score, imply a lower score assigned to it. In the latter case, crops of the object targeted by the action (\textit{e.g.} wine glass) are not meaningful without crops of the instruments performing the action (\textit{e.g.} wine bottle) in order to infer the action (\textit{e.g.} fill a wine glass).

    \item[\textit{(iii)}] Figures \ref{fig:ranking_neg}\subref{fig:rank4_neg},\subref{fig:rank5_neg},\subref{fig:rank6_neg} show examples of the HOI classes that experience a larger performance drop when applying Black Other Objects for object specification. In all examples there are several people performing the same action. The drop in performance may stem from the removal of additional examples, which may help the MLLM reinforce the certainty of interaction and better understand the scene. The opposite effect happens when using this prompting strategy with several individuals performing different actions on the same object, as explained in observation \textit{(vii)}.

    \item[\textit{(iv)}] Figure \ref{fig:HOI_difference} shows the number of HOI classes whose performance either improves or worsens when introducing each prompting method, compared to the baseline of Qwen2-VL 2B. It is observed that Objects as Context and Black Other Objects, along with Crop, are the only strategies able to improve the performance for a majority of HOIs.

    \item[\textit{(v)}] Figure \ref{fig:worst_classes} highlights the worst-performing HOI classes across all object specification prompting methods. For the classes \textit{Hop on horse} and \textit{Jump boat}, Objects as Context slightly improves performance relative to both the baseline and the average of the prompting methods. Similarly, for \textit{Exit boat} and \textit{Open umbrella}, Black Other Objects proves beneficial. Overall, performance on these classes remains very low, underscoring the inherent difficulty of the task for any prompting strategy.

    \item[\textit{(vi)}] Figure \ref{fig:obj_count} shows the performance according to the number of objects detected that belong to the target class. As illustrated, the higher the number of objects detected of the same class, the more likely the MLLM will be confused. To avoid specification confusion, MLLMs should be able to correctly understand prompting techniques to focus on the specified object instance.

    \item[\textit{(vii)}] Figures \ref{fig:use_cases}\subref{fig:use1},\subref{fig:use2},\subref{fig:use3},\subref{fig:use4} show examples of images and HOI classes that Objects as Context predict successfully but all other prompting methods fail to detect. Objects as Context may provide the MLLM with richer contextual cues on the specified objects by providing crops of the objects and people. These crops might be especially important when objects or people are either small (Figures \ref{fig:use_cases}\subref{fig:use1},\subref{fig:use2}) or partially occluded (Figure \ref{fig:use_cases}\subref{fig:use3}), since their visual tokens might be enhanced.

    \item[\textit{(viii)}] Figures \ref{fig:use_cases}\subref{fig:use5},\subref{fig:use6} show examples of images and HOI classes, along with their representation after applying the visual prompting strategy, that Black Other Objects predict successfully while all other prompting methods tested fail to detect. In both cases, there are many people interacting differently with the same object class (bicycle for Figure \ref{fig:use_cases}\subref{fig:use5} and elephant for Figure \ref{fig:use_cases}\subref{fig:use6}), which may confuse the MLLM. Black Other Objects is able to erase misleading information from the scene while keeping the relevant context, thus helping the MLLM direct its attention to the correct area in the image.
\end{enumerate}

In summary, isolating detection errors through this evaluation protocol not only clarifies the impact of visual prompting on object specification but also provides interpretable insights into how MLLMs handle this challenge, paving the way for more reliable zero-shot approaches to open-world HOI.

\subsection{Ablation studies} \label{sec:ablation_studies}
To better understand the effects and robustness of the different components of the proposed HOI recognition pipeline, we perform a series of ablation studies on the HICO dataset using mainly Qwen2-VL 2B. These experiments analyze how different components and hyperparameters influence HOI recognition performance and evaluation computational cost. Specifically, we study the computation of the \textit{no\_interaction} score, the impact of the object detector model, the effect of object-class filtering through the threshold $\tau$, and the role of the context-selection parameters $k_1$ and $k_2$.

\subsubsection{No-interaction score}
We analyze the behavior of the \textit{no\_interaction} computation by splitting the 600 HICO classes into 520 interacting classes and 80 \textit{no\_interaction} classes. The interacting subset reaches 57.87 mAP, whereas the \textit{no\_interaction} subset obtains 9.32 mAP. This gap indicates that the product-based score in Equation \ref{eq:no_interaction} accumulates uncertainty for this class: although it makes the score sounder, ambiguous positive scores across many interactions can quickly reduce the final \textit{no\_interaction} confidence. However, treating \textit{no\_interaction} as an independent class with its own questions could lead to inconsistent outputs, such as all scores being low, including \textit{no\_interaction}, or assigning high scores to \textit{no\_interaction} and interactive classes simultaneously.

\subsubsection{Object Detector}
Since our pipeline relies on an external object detector as the entry point, we analyze how sensitive the overall HOI recognition performance is to the choice of object detector backbone. Table \ref{tab:od_model} reports results for four established detectors, YOLOX-x (\cite{ge2021yolox}), YOLO11-x (\cite{khanam2024yolov11}), Faster R-CNN (\cite{ren2015faster}), and DINO (\cite{zhang2022dino}), evaluated with MiniCPM-V 1 and Qwen2-VL 2B. Performance differences across detectors are modest: DINO achieves the highest mAP (55.49 with Qwen2-VL 2B), but at the cost of retaining over 350k queries, which substantially increases evaluation overhead. YOLO11-x offers the most favorable trade-off, reaching competitive mAPs of 48.20 (MiniCPM-V 1) and 51.39 (Qwen2-VL 2B) while minimizing the number of retained queries ($\approx$ 252k). For this reason, we adopt YOLO11-x as the default detector in our pipeline.

\begin{table}[ht]
    \centering
    \begin{tabular}{@{} l c  c  c @{}}
        \toprule
        Object Detector & MiniCPM-V 1 & Qwen2-VL 2B & \#Questions\\
        \midrule
        YOLOX-x (\cite{ge2021yolox}) & 48.59 & 52.45 & 275,746 \\
        YOLO11-x (\cite{khanam2024yolov11}) & 48.20 & 51.39 & \textbf{251,593} \\
        Faster R-CNN (\cite{ren2015faster}) & 48.31 & 52.28 & 351,466 \\
        DINO (\cite{zhang2022dino}) & \textbf{51.27} & \textbf{55.49} & 352,653 \\
        \bottomrule
    \end{tabular}
    \caption{Comparison of object detectors. Evaluation is conducted on HICO. \# Questions denotes the number of retained queries after the object detector filtering step. All models are trained on MS-COCO.}
    \label{tab:od_model}
\end{table}

To further illustrate detector robustness, we include in the Appendix an extended comparison of YOLO11 model sizes and an exploratory experiment where the MLLM itself is used as an object recognizer. Overall, these results show that our benchmark setup is not overly sensitive to the specific choice of detector, and that the main trends reported in Sections \ref{sec:hoi_recognition_res}–\ref{sec:hoi_detection_gt} hold across reasonable detection backbones.

\subsubsection{Class-wise object detection threshold}
We next analyze the effect of the object-class filtering stage, which determines which HOI categories are questioned to the MLLM. This experiment aims to quantify how much the pipeline depends on the object detector-based filtering step. Table \ref{tab:tau_threshold} evaluates different values of the object class score threshold $\tau$. Lowering $\tau$ increases the number of retained object classes, which consequently increases the number of questions evaluated by the MLLM. If the pipeline relied strongly on this filtering stage, retaining more object classes would be expected to produce substantial performance gains. We select $\tau = 0.10$ as the default operating point because it provides a favorable balance between recognition accuracy and computational cost. More importantly, the relatively small performance variation across thresholds shows that the object-class filtering stage does not dominate the final recognition accuracy, indicating that the proposed pipeline remains robust even when the number of retained object classes changes considerably.

\begin{table}[ht]
    \centering
    \begin{tabular}{@{} l c c c @{}}
        \toprule
        $\tau$ & MiniCPM-V 1 & Qwen2-VL 2B & \#Questions \\
        \midrule
        0.00 & \textbf{50.98} & \textbf{53.59} & 990,910 \\
        0.05 & 49.33 & 52.61 & 273,668 \\
        0.10 & 48.20 & 51.39 & 251,593 \\
        0.25 & 47.92 & 51.67 & 223,304 \\
        0.40 & 47.06 & 50.82 & 206,492 \\
        0.50 & 46.23 & 49.63 & \textbf{195,441} \\
        \bottomrule
    \end{tabular}
    \caption{Effect of varying class score thresholds in the pipeline. Evaluation conducted on HICO.}
    \label{tab:tau_threshold}
\end{table}

\subsubsection{Number of objects and people retained}
In Objects as Context, hyperparamters $k_1$ and $k_2$ control the maximum number of objects and people provided to the MLLM as context. Objects are filtered by target class and ranked by detection confidence, while people are ranked by bounding box size, prioritizing larger instances as they should represent more relevant people in the scene. Tables \ref{tab:num_objects} and \ref{tab:num_people} show that there is no great difference between the values, but $k_1=10$ and $k_2=3$ are selected because of their slightly superior results.

\begin{table}[ht]
    \centering
    \begin{minipage}[b]{0.45\linewidth}
        \centering
        \begin{tabular}{@{} l c @{}}
            \toprule
            \# Objects ($k_1$) & mAP \\
            \midrule
            1   & 53.35 \\
            5   & 53.45 \\
            10  & \textbf{53.50}\\
            20  & 53.47 \\
            50  & 53.47 \\
            \bottomrule
        \end{tabular}
        \caption{Effect of varying the number of objects kept for Objects as Context. Evaluation conducted on HICO using Qwen2-VL 2B.}
        \label{tab:num_objects}
    \end{minipage}
    \hfill
    \begin{minipage}[b]{0.45\linewidth}
        \centering
        \begin{tabular}{@{} l c @{}}
            \toprule
            \# People ($k_2$) & mAP \\
            \midrule
            1  & 53.45 \\
            3  & \textbf{53.50} \\
            5  & 53.44 \\
            10 & 53.47 \\
            \bottomrule
        \end{tabular}
        \caption{Effect of varying the number of people kept for Objects as Context. Evaluation conducted on HICO using Qwen2-VL 2B.}
        \label{tab:num_people}
    \end{minipage}
\end{table}

\section{Conclusions, Limitations, and Future Work} \label{sec:conclusions}
We presented the first benchmark and prompting taxonomy for \emph{pure zero-shot} HOI recognition and detection using MLLMs and proposed two effective visual prompting strategies, Objects as Context and Black Other Objects, which improve performance by guiding model attention and reducing distractions. Evaluations on the constructed Q\&A datasets established a strong \emph{pure zero-shot} baseline, achieving zero-shot state-of-the-art results (53.50 mAP on HICO, 23.69 mAP on HICO-DET, 32.27 mAP on V-COCO) without HOI-specific fine-tuning.

While these results are promising, several limitations should be considered. First, the reported evaluation performance inherits the biases of the underlying HOI datasets, including their annotation errors and fixed interaction categories. These biases may affect the evaluation protocol and the interpretation of the reported metrics. However, this limitation does not affect the model's generalization capabilities, since it is evaluated in a \emph{pure zero-shot} setting.

Second, our pipeline relies on pre-trained models, namely the external object detector and the MLLMs used for interaction reasoning, whose capabilities and errors may limit the overall performance. Because the detector is restricted to the object categories learned during training, in our case COCO classes, the complete pipeline is not fully open-world and may miss objects outside this vocabulary. This restriction does not limit the underlying HOI recognition capabilities of the MLLM, since the detector is used only to filter the questions evaluated for efficiency and can be removed at deployment. For HOI detection, future work could instead leverage the grounding capabilities of recent MLLMs to localize open-vocabulary objects directly.

Third, although the proposed pipeline incorporates several practical optimizations to reduce computational and memory requirements, such as object class filtering, evaluating MLLMs for HOI datasets remains computationally demanding due to the large number of image--question pairs that must be processed. These design choices introduce an inherent trade-off between computational efficiency and evaluation completeness. Developing more efficient evaluation strategies that preserve performance while reducing inference cost remains an important direction for future research.

Fourth, \textit{no\_interaction} labels remain challenging. Future work should improve their scoring while balancing output consistency with accurate estimation of interaction absence.

Overall, this benchmark enables systematic evaluation of MLLMs on open-world HOI and provides a basis for extending the proposed methodology to other multi-label problems, particularly those characterized by long-tail class distributions, where \emph{pure zero-shot} inference may reduce reliance on imbalanced annotations.

\section*{Ethical considerations}
HOI datasets such as HICO, HICO-DET, and V-COCO often include images sourced from the internet that depict people in everyday activities. While our work focuses on technical benchmarking, methods developed here could potentially be applied in sensitive domains such as surveillance or security. We emphasize that such applications raise privacy and ethical concerns, and we encourage future work to carefully consider safeguards and responsible deployment.

\section*{Acknowledgements}
This work has been partially supported by the Spanish project PID2025-170489NB-I00 and by ICREA under the ICREA Academia programme.

\bibliography{main}

\newpage
\begin{appendices}
\counterwithin{table}{section}
\counterwithin{figure}{section}
For a better understanding of the main paper, we provide additional details in this appendix, including:

\begin{itemize}
    \item Section \ref{sec:ann_mcot} shows the prompts used to implement Multimodal Chain-of-Thought.
    \item Section \ref{sec:ann_caption} shows the prompts used to implement Caption as Context.
    \item Section \ref{sec:ann_ensemble} shows the procedure to generate prompts for Ensemble of Prompts.
    \item Section \ref{sec:ann_textual} shows examples of textual prompts used for each object specification prompting strategy.
    \item Section \ref{ann:hyperparameters} further lists the hyperparameters used.
    \item Section \ref{sec:od_size} studies the effect of the size of the object detector on our pipeline.
\end{itemize}

\section{Multimodal Chain-of-Thought templates} \label{sec:ann_mcot}
The reasoning generation step adds the sentence ``Think step-by-step.'' (\cite{kojima2022large}) at the end of the question. The final question prompt is \texttt{Context: <reasoning>. Question: <Question>. Answer with yes or no.}.

\section{Caption as Context templates} \label{sec:ann_caption}
The caption generation step uses ``Describe the image in detail.''. The final question prompt is \texttt{Context: <caption>. Question: <Question>. Answer with yes or no.}.

\section{Ensemble of Prompts generation} \label{sec:ann_ensemble}
We use the action synonyms provided in the action description of the HICO (\cite{chao2015hico}) dataset to build new prompt templates for each HOI class. The generation of questions is shown in Figure \ref{fig:ensemble_generation}. The different templates for \textit{person\_option} are adapted from \cite{gupta2024exploring}.

\begin{tcolorbox}[colback=black!5, colframe=black, title={Creation of questions}]
\begin{lstlisting}[escapeinside={(*@}{@*)}]
{(*@\textcolor{green}{person\_option}@*)} {(*@\textcolor{blue}{action}@*)} a {(*@\textcolor{red}{object}@*)}? Answer yes or no.

Where:
(*@\textcolor{green}{person\_option}@*) in ["Is there anybody", 
                "Are the people in the image", 
                "Are the individuals in the image",
                "Are the humans in the image",
                "Are the people", 
                "Are the individuals", 
                "Are the humans"]
(*@\textcolor{blue}{action}@*) in ["carrying", "transporting"]
(*@\textcolor{red}{object}@*) is "bike"
\end{lstlisting}
\end{tcolorbox}
\noindent\begin{minipage}{\textwidth}
\captionof{figure}{Prompt generation for ensemble of prompts. A prompt is generated by substituting each part of the main question (top part) with combinations of person\_option, action, and object. Each object-action pair has its own set of possible action synonyms.}\label{fig:ensemble_generation}
\end{minipage}

In the cases where the desired number of questions is lower than the possible number of questions, a subset is retrieved at random. In the opposite case, when the ensemble number is higher than the possible number of questions, all are retrieved, and the first ones are computed more than once. Notice that the number of questions differs by the specific action since each verb has a different number of synonyms.

\section{Object specification prompting strategy textual templates} \label{sec:ann_textual}
Following the visual example in Figure \ref{fig:visual_prompt_examples}, the question for the HOI \textit{carry backpack} for each object specification strategy tested is:

\begin{itemize}
    \item Baseline: \texttt{Is there anybody carrying a backpack? Answer with yes or no.}
    \item Objects as Context: \texttt{Is there anybody carrying a backpack? Answer with yes or no.}
    \item Overlay Bounding Boxes: \texttt{Is the person in the blue box carrying the backpack in the red box? Answer yes or no.}
    \item Overlay Elliptic Markers: \texttt{Is the person within the blue marker carrying the backpack within the red marker? Answer yes or no.}
    \item Blur the Background: \texttt{Is the foreground person carrying a backpack? Answer yes or no.}
    \item Crop the relevant area: \texttt{Is there anybody carrying a backpack? Answer with yes or no.}
    \item Black Other Objects: \texttt{Is there anybody carrying a backpack? Answer with yes or no.}
    \item Gray the Background: \texttt{Is the colored person carrying the colored backpack? Answer yes or no.}
    \item Textual Center Coordinates: \texttt{Is the person centered at (209, 368) carrying the backpack centered at (209, 320)? Answer yes or no.}
    \item Textual Boundary Coordinates: \texttt{Is the person within image coordinates (168, 243), (168, 493), (250, 243) and (250, 493) carrying the backpack within image coordinates (177, 278), (177, 363), (241, 278) and (241, 363)? Answer yes or no.}
\end{itemize}

\subsection{Thresholded accuracy analysis}
\label{sec:appendix_accuracy}
One of the objectives of this work is to provide a benchmark for \emph{pure zero-shot} HOI recognition that is directly comparable with existing methods. Therefore, the main paper reports results using mean Average Precision (mAP), the standard evaluation metric for HOI recognition. Since some prompting strategies tend to produce more confident predictions, we additionally analyze their binary classification performance after thresholding.

For each HOI query, the predicted probability is converted into a binary prediction using a threshold of $0.5$. Predictions with a score greater than or equal to $0.5$ are assigned the positive label, while the remaining predictions are assigned the negative label. The final accuracy is then computed as the percentage of correctly classified HOI queries over the entire evaluation set.

\begin{table}[ht]
    \centering
    \begin{tabular}{@{}lccc@{}}
        \toprule
        Strategy & MiniCPM-V 1 & Qwen2-VL 2B &  LLaVA-OneVision 7B \\
        \midrule
        Baseline & 98.88 & 99.10 & 98.68 \\
        MCoT & 99.15 & 98.99 & - \\
        Caption as Context & 99.03 & 99.21 & - \\
        Objects as Context & 98.67 & 99.16 & 99.51 \\
        \bottomrule
    \end{tabular}
    \caption{Thresholded accuracy (\%) on HICO. Binary predictions are obtained by thresholding the predicted probabilities at $0.5$.}
    \label{tab:threshold_accuracy}
\end{table}

Table \ref{tab:threshold_accuracy} reports the corresponding thresholded accuracy. The Multi-step with Generated Knowledge strategies achieve accuracy comparable to or slightly higher than the baseline, despite obtaining lower mAP in the main evaluation. This indicates that these strategies tend to produce more decisive predictions, improving the final classification after thresholding, while yielding score distributions that are less suitable for the ranking-based mAP metric. In contrast, Objects as Context provides the largest gains under mAP while maintaining a comparable thresholded accuracy, suggesting that its main benefit lies in improving the calibration and ranking of prediction scores rather than the binary decisions themselves.

\color{black}

\section{List of hyperparameters} \label{ann:hyperparameters}
\begin{table}[ht]
    \centering
    \begin{tabular}{@{} l l @{}}
        \toprule
        Hyperparameter & Value \\
        \midrule
        Attention implementation & Flash Attention 2 from \cite{dao2023flashattention} \\
        Generation strategy & Greedy \\
        Maximum number of new tokens & 1 \\
        Maximum number of output tokens for generated knowledge & 256 \\
        Minimum size of cropped objects\protect\footnotemark & 28 per side \\
        \bottomrule
    \end{tabular}
    \caption{List of hyperparameters for MLLMs in inference mode.}
\end{table}
\footnotetext{Only apply to Qwen2-VL and LLaVA-OneVision models.}

\section{Object detector size} \label{sec:od_size}
Table \ref{tab:od_size} shows the results of using different model sizes within the YOLO11 family. Performance scales smoothly with detector size up to the \textit{large} variant. In addition, we report an approach that uses MLLMs as object recognizers based on the proposed classification score, thus removing the necessity of a pre-trained object detector. This approach directly queries the presence or absence of objects without localization, prompting one question per object class (\textit{e.g.}, 80 per image for MS-COCO). However, this approach results in a very low performance for Qwen2-VL 2B, possibly because leveraging the same MLLM for both object recognition and HOI recognition propagates errors between tasks, highlighting the necessity of a dedicated detector.

\begin{table}[ht]
    \centering
    \begin{tabular}{@{} l c c c @{}}
        \toprule
        Model & MiniCPM-V 1 & Qwen2-VL 2B & \#Questions \\
        \midrule
        nano & 43.37 & 47.16 & \textbf{240,581} \\
        small & 45.86 & 49.73 & 249,280 \\
        medium & 47.28 & 51.15 & 251,102 \\
        large & \textbf{48.55} & \textbf{52.32} & 252,084 \\
        extra-large & 48.20 & 51.39 & 251,593 \\
        MLLM \scriptsize (Qwen2-VL 2B) & 50.96 & 14.45 & 264,782 \\
        \bottomrule
    \end{tabular}
    \caption{Performance across YOLO11 model sizes. Evaluation conducted on HICO. All detectors are trained on MS-COCO.}
    \label{tab:od_size}
\end{table}
\end{appendices}

\end{document}